%% file: neurips_main.tex
\documentclass{article}

\RequirePackage[numbers,sort&compress]{natbib}

\usepackage[final]{neurips_2024}

\usepackage[utf8]{inputenc} 
\usepackage[T1]{fontenc}    
\usepackage{hyperref}       
\usepackage{url}            
\usepackage{booktabs}       
\usepackage{amsfonts}       
\usepackage{nicefrac}       
\usepackage{microtype}      
\usepackage{xcolor}         

\usepackage{wrapfig}
\usepackage{microtype}
\usepackage{graphicx}
\usepackage{subfigure}
\usepackage{booktabs} 
\usepackage{algorithm}
\usepackage{algorithmic}

\usepackage{amsmath}
\usepackage{amssymb}
\usepackage{mathtools}
\usepackage{amsthm}

\usepackage[capitalize,noabbrev]{cleveref}

\theoremstyle{plain}

\theoremstyle{definition}

\theoremstyle{remark}

\usepackage[textsize=tiny]{todonotes}

\usepackage{tcolorbox}
\usepackage{multirow}
\usepackage{MnSymbol}
\usepackage{physics}
\usepackage{bm}
\definecolor{LightGray}{gray}{0.9}

\newcommand{\mr}{{\textbf{MR}}}
\newcommand{\dm}{{\bm{$\Delta m\%$}}}

\definecolor{contextcolor}{RGB}{31,160,171}
\newcommand{\best}[1]{{\textbf{\textcolor{jiayiblue}{#1}}}}

\newcommand{\stl}{\textsc{STL}}
\newcommand{\ls}{\textsc{LS}}
\newcommand{\si}{\textsc{SI}}
\newcommand{\rlw}{\textsc{RLW}}
\newcommand{\dwa}{\textsc{DWA}}
\newcommand{\uw}{\textsc{UW}}
\newcommand{\famo}{\textsc{FAMO}}
\newcommand{\graddrop}{\textsc{GradDrop}}

\newcommand{\mgda}{\textsc{MGDA}}
\newcommand{\pcgrad}{\textsc{PCGrad}}
\newcommand{\imtlg}{\textsc{IMTL-G}}
\newcommand{\cagrad}{\textsc{CAGrad}}
\newcommand{\nashmtl}{\textsc{NashMTL}}
\newcommand{\ours}{\textbf{\textit{GO4Align}}}

\usepackage{tikz}
\usepackage{pifont}
\newcommand{\ymark}{\ding{51}}%

\definecolor{updatecolor}{rgb}{0,0,1}

\definecolor{camera_ready}{rgb}{0,0,0}

\definecolor{jiayiblue}{rgb}{0.1,0.2,0.7}
\definecolor{myred}{rgb}{0.858, 0.1, 0.1}
\definecolor{myyellow}{rgb}{0.858, 0.6, 0.1}
\definecolor{mydarkred}{rgb}{0.69,0.0,0.1098}

\definecolor{commentcolor}{RGB}{110,154,155}   

\usepackage{hyperref}
\hypersetup{
    colorlinks=true,
    linkcolor=blue,
    citecolor=green,      
    urlcolor=cyan
    }

\title{\ours{}: Group Optimization for \\ Multi-Task Alignment}

\author{Jiayi Shen$^{1}$,  Qi (Cheems) Wang$^{2}$, Zehao Xiao$^{1}$\thanks{Correspondence to: Zehao Xiao <zxiao4ai@gmail.com>.}~, Nanne Van Noord$^{1}$, Marcel Worring$^{1}$\\
$^1$University of Amsterdam, Amsterdam, the Netherlands\\ 
$^2$Department of Automation, Tsinghua University, Beijing, China
}

\begin{document}
\maketitle
\input{section/0_abstract}
\input{section/1_intro}

\input{section/2_methods}

\input{section/3_relatedworks}

\input{section/4_experiments}

\input{section/5_conclusions}


{
\small
\bibliography{neurips_main}
\bibliographystyle{unsrtnat}
}

\input{section/6_appendix}
\input{section/7_checklist}

\end{document}

%% file: section/0_abstract.tex
\begin{abstract}

This paper proposes \ours{}, a multi-task optimization approach that tackles task imbalance by explicitly aligning the optimization across tasks. To achieve this, we design an adaptive group risk minimization strategy, comprising two techniques in implementation: (i) dynamical group assignment, which clusters similar tasks based on task interactions; (ii) risk-guided group indicators, which exploit consistent task correlations with risk information from previous iterations. Comprehensive experimental results on diverse benchmarks demonstrate our method's performance superiority with even lower computational costs.
\end{abstract}

%% file: section/1_intro.tex
\section{Introduction}
\label{sec: introduction}

\input{fig/fig1}

Multi-task learning is a promising paradigm for handling several tasks simultaneously using a unified architecture.
It can achieve data efficiency, improve generalization, and reduce computation costs compared with addressing each task individually~\cite{vandenhende2021multi}.
Due to these benefits, there is a growing surge of applications with multi-task learning in several domains, \textit{e.g.}, natural language processing~\cite{chen2021multi, pilault2020conditionally, liu2017adversarial}, computer vision~\cite{vandenhende2021multi, kendall2018multi, liu2019end} and reinforcement learning~\cite{espeholt2018impala, sodhani2021multi}. 
The crux of multi-task learning is to enable positive transfer among tasks while avoiding negative transfer, which usually exists among irrelevant tasks~\cite{xin2022current, liu2022auto, standley2020tasks}.

\textbf{Existing Challenges:}
In avoiding the negative transfer, numerous multi-task optimization (MTO) methods~\cite{yu2020gradient, kendall2018multi, navon2022multi, xin2022current, liu2023famo} have emerged and attracted rising attention in recent years. 
A lasting concern in MTO is the \textit{task imbalance issue}.
It describes a phenomenon where some tasks are severely under-optimized~\cite{vandenhende2021multi}, which can lead to worse overall performance with larger convergence differences across tasks.

To deal with the task imbalance issue, scaling methods are proposed for MTO.
According to differences in scaling manipulations, we roughly divide MTO methods into \textit{gradient-oriented}~\cite{yu2020gradient, senushkin2023independent, liu2021conflict, chen2020just, sener2018multi} and \textit{loss-oriented}~\cite{kendall2018multi, liu2019end, kurin2022defense, lin2021reasonable}.
The former tends to exhibit impressive results at the expense of higher computational or memory requirements during training time due to the assessment of per-task gradients~\cite{kurin2022defense}.\footnote{In MTO, computational efficiency refers to training time efficiency.}
In contrast, the latter preserves training-time efficiency but usually suffers from unsatisfactory overall performance. 
As shown in Fig.~\ref{fig: fig1}, most existing methods cannot simultaneously achieve superior performance and computational efficiency. 

\textbf{Proposed Solution:}
To improve the overall performance and maintain computational and memory efficiency, 
we propose \textbf{\underline{G}}roup \textbf{\underline{O}}ptimization \textbf{\underline{for}} multi-task \textbf{\underline{Align}}ment (\ours{}), a novel and effective loss-oriented method in MTO.
As shown in Fig.~{\ref{fig: fig2}}, this work identifies \textit{multi-task alignment} as a crucial factor in solving task imbalance, which means learning progress across tasks should synchronously achieve superior performance over all tasks. The proposed model dynamically aligns learning progress across tasks by exploiting group-based task interactions for multi-task empirical risk minimization.
The rationale behind this is that groupings can implicitly capture task correlations for more effective multi-task alignment and thus help multi-task optimizers benefit from positive interactions among relevant tasks.
The primary contribution is two-fold:
\vspace{-2mm}
\begin{itemize}
    \item As a new member of the loss-oriented MTO branch, \ours{} recasts the task imbalance issue to a bi-level optimization problem, yielding an adaptive group risk minimization principle for MTO. Such a principle allocates weights over task losses at a group level to achieve learning progress alignment among relevant tasks.
    \item We develop a heuristic optimization pipeline in \ours{} to tractably achieve the principle, involving \textit{dynamical group assignment} and \textit{risk-guided group indicators}. 
    The pipeline incorporates beneficial task interactions into the group assignments and exploits task correlations for multi-task alignment, improving overall multi-task performance. 
\end{itemize}
\vspace{-2mm}
Experimental results show that our approach can outperform existing state-of-the-art baselines in extensive benchmarks. Moreover, it does not sacrifice computational efficiency.

%% file: fig/fig1.tex
\begin{wrapfigure}{r}{0.5\textwidth}
\centering
\vspace{-15pt}
\includegraphics[width=1.0\linewidth]{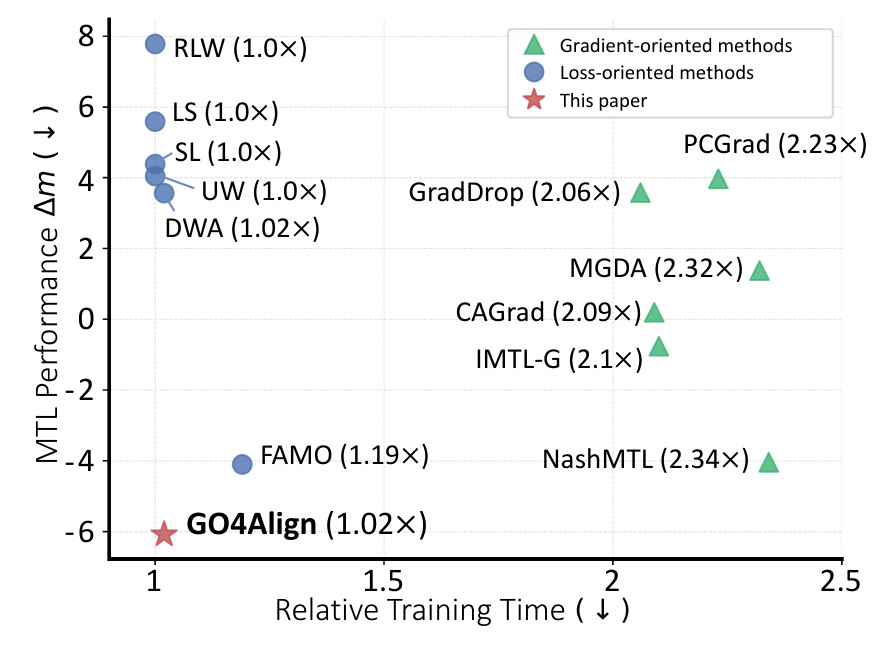}
\vspace{-23pt}
\caption{\textbf{Performance and computational efficiency evaluation for MTO methods evaluated on}~\texttt{NYUv2}. Each method's training time is relative to a baseline method, which minimizes the sum of task-specific empirical risks. Left-bottom marks comprehensive optimal results.}
\vspace{-15pt}
\label{fig: fig1}
\end{wrapfigure}

%% file: section/2_methods.tex
\section{Preliminary}

\noindent\textbf{Notations.} 
This work considers a multi-task problem over an input space $\mathcal{X}$ and a collection of task target spaces $\{\mathcal{Y}^m\}_{m=1}^M$, where $M \geq 2$
denotes the number of tasks. A composite dataset for multi-task learning is $\{ ({\bm{x}}_n, y_n^1, ..., y_n^M)\}_{n=1}^{N}$, where $N$ is the number of training samples. Let $\bm{\theta}^{s}$ and $\bm{\theta}^{m}$ respectively be the shared and task-specific parameters in a given multi-task model, thus we have a parametric hypothesis class for the $m$-th task as 
$f({\bm{x}}_n; \bm{\theta}^{s}, \bm{\theta}^{m}): \mathcal{X} \rightarrow \mathcal{Y}^m$. Then the empirical risk for the $m$-th task can be written as $\hat{\mathcal{L}}^m( \bm{\theta}^{s}, \bm{\theta}^{m}) = \frac{1}{N} \sum_{n=1}^{N} \ell^m(f({\bm{x}}_n; \bm{\theta}^{s}, \bm{\theta}^{m}), y_n^m)$ , where $\ell^m(\cdot, \cdot): \mathcal{Y}^m \times \mathcal{Y}^m \rightarrow \mathbb{R}_{+}$ denotes the task-specific loss function. The ultimate goal of MTO is to achieve superior performance over all tasks. 

\noindent\textbf{Scale Empirical Risk Minimization (Scale-ERM).} As preliminary, we recap a representative and related strategy in MTO, Scale-ERM, through the lens of risk minimization.
Scale-ERM introduces a task-specific weight $\lambda^m \geq 0$ to scale the corresponding empirical risk. For conciseness, this principle is formulated using vector notations. Here, $\bm{\lambda} = \left[\lambda^1, \lambda^2, \cdots, \lambda^M \right]^{\top} \in \mathbb{R}^M$ represents a $M$-dimensional vector comprising all task-specific weights. And $\hat{\bm{L}}(\bm{\theta}) = \left[\hat{\mathcal{L}}^1(\bm{\theta}^{s}, \bm{\theta}^{1}), \hat{\mathcal{L}}^2(\bm{\theta}^{s}, \bm{\theta}^{2}), \cdots, \hat{\mathcal{L}}^M(\bm{\theta}^{s}, \bm{\theta}^{M})\right]^{\top} \in \mathbb{R}^M$ denotes the corresponding vector of empirical risks, where $\bm{\theta} = \{\bm{\theta}^{s},\bm{\theta}^{1}, \bm{\theta}^{2}, \cdots, \bm{\theta}^{M}\}$ represents all learnable parameters in a multi-task backbone network. Thus, we obtain the objective of Scale-ERM as follows:
\begin{equation}
\begin{aligned}
    \operatorname*{min}_{\bm{\theta},\bm{\lambda}}  {\bm{\lambda}}^{\top} 
    \hat{\bm{L}}({\bm{\theta}})+{\Omega}(\bm{\lambda}),
\end{aligned}
\label{eq: scale-ERM}
\end{equation}
where ${\Omega}(\bm{\lambda})$ is a regularization term over task weights, designed to prevent the rapid collapse of these weights to zero, as discussed in~\citet{kendall2018multi}. When all task weights are the same in scale, Scale-ERM will degenerate to the most simple strategy in MTO, where each task is treated equally during the joint training.  

In Scale-ERM, each task weight controls task-specific learning progress, either by adapting the task-specific weights with loss information (loss-oriented) or by operating on task-specific gradients (gradient-oriented). 
As previously indicated, there still remains a research gap in MTO to improve multi-task performance without affecting computational efficiency.
\section{Methodology}

In resolving the task imbalance issue effectively and efficiently, we develop \ours{} in this section. 
Our approach relies on grouping-based task interactions to align learning progress across tasks. As a new member of the loss-oriented branch, \ours{} holds the advantage of computational efficiency without the requirements of per-task gradients.

\input{fig/demo2}
\noindent \textbf{Motivation of Multi-Task Alignment.} Empirically, we can observe that the overall performance is worse when there is a larger \textit{convergence difference} between tasks. The convergence difference is measured by the standard deviation of the task-specific epoch numbers to reach convergence.
As shown in Fig.~\ref{fig: fig2}, 
\uw{}~\cite{kendall2018multi} underperforms \famo{}~\cite{liu2023famo} in terms of a overall MTL metric \dm~(lower is better); while \uw{} has a larger convergence difference than \famo{}. Intuitively, a larger convergence difference means that per-task training dynamics are more asynchronous, usually leading to worse overall performance.

To perform MTO, we consider aligning tasks with group information in the multi-task risk minimization. We first present an adaptive group risk minimization principle for MTO, which targets the alignment of tasks' learning progress from grouping-based task interactions. 
Then, in tractable problem-solving, we decompose the whole optimization process into two entangled phases: \textit{(i) dynamical group assignment} and \textit{(ii) risk-guided group indicators}.
The pseudo-code of \ours{} is provided in Appendix~\ref{sec: algorithm}.

\subsection{Adaptive Group Risk Minimization Principle}
\label{sec: AGRM}
Recent advances~\cite{fifty2021efficiently, kang2011learning} have explored incorporating multi-task grouping in feature sharing and revealed its benefit of aligning the learning progress through task interactions.
Nevertheless, their grouping mechanisms ignore monitoring the learning progress, \textit{e.g.}, failure to capture variations in loss scales among tasks, weakening the effectiveness of multi-task alignment.

As a result, we design a new grouping mechanism for MTO with \textit{task-specific learning dynamics}, which directly impacts the \textit{convergence behaviors in optimization}. 
This induces the adaptive group risk minimization principle suitable for multi-task alignment.
The hypothesis is that the dynamical grouping tends to implicitly exploit task correlations~\cite{fifty2021efficiently} and encourages beneficial task interactions from empirical risk information along the learning progress.
Meanwhile, such a principle retains computational efficiency as it avoids the computations of per-task gradients.  

\noindent\textbf{Adaptive Group Risk Minimization (AGRM).} 
We first achieve beneficial task interactions by producing task weights with a grouping mechanism, which is adaptive to various loss scales and their learning dynamics over time. 
Then, we recast the task imbalance issue with the grouping mechanism into a bi-level optimization problem: 
(i)~In the \textit{lower-level} optimization, the model aims to cluster $M$ tasks of interest into $K$ groups. This implicitly exploits task correlations, where similar tasks should be clustered into one group, yielding more beneficial task interactions.
With group assignments, group weights are designed to conduct learning progress alignment at the group level.
\textcolor{camera_ready}{
The group weights in the multi-task objective are motivated as an extension of the observation that similar tasks benefit greatly from training together through parameter sharing~\cite{fifty2021efficiently}.
}
(ii)~In the \textit{upper-level} optimization, the proposed principle updates model parameters from the grouped empirical risks, which implicitly relies on the lower-level optimized results. 
We illustrate \ours{}'s bi-level optimization with grouping-based task interactions in Fig.~\ref{fig: graphmodel}.

\input{fig/GO4Align}

Given $K$ as the number of groups, we denote the assignment matrix as $\bm{\mathcal{G}}_t \in \mathbb{R}^{K \times M}$, where $\mathcal{G}_t(k, m)$ equals $1$ if the $k$-th group contains the $m$-th task and $0$ otherwise. 
Note that $\bm{\mathcal{G}}_t$ is updated in the optimization, with $t$-th indexing iteration step. 
The group number generally is smaller than the task number, \textit{e.g.},  $1 < K \leq M$.
To balance different groups, we place weights over groups $\bm{\omega}_t = \left[\omega_t^1, \omega_t^2, \cdots, \omega_t^K \right]^{\top} \in \mathbb{R}^K$,  where $\omega_t^k \geq 0$ is specific to the $k$-th group at the $t$-th iteration. 
Formally, we formulate the bi-level optimization problem as:
\begin{equation}
\begin{aligned}
        \operatorname*{min}_{\bm{\theta}} 
        {{{\bm{\omega}}_t}^{\top}} {\bm{\mathcal{G}}}_t \hat{\bm{L}}(\bm{\theta}) 
        ~~~~ \textit{s.t.}~\{{{\bm{\omega}}_t},{\bm{\mathcal{G}}}_t\} = \arg\operatorname*{min}_{\bm{\omega},\bm{\mathcal{G}}}\bm{J}(\bm{\omega},\bm{\mathcal{G}}; \bm{\theta}_{t}),
\end{aligned}
\label{eq: group-ERM}
\end{equation} 
where $\bm{\omega}_t$ and $\bm{\mathcal{G}}_t$ reflect adaptive group information in the lower-level optimization. 
$\bm{J}(\bm{\omega},\bm{\mathcal{G}}; \bm{\theta}_t)$ is the corresponding optimization objective for aligning the learning progress across tasks at the group levels, which is further explained in Sec.~\ref{sec: dynamic_group_assignment}.

To be specific, we perform the bi-level optimization in Eq.~(\ref{eq: group-ERM}) according to the following steps. For the $t$-th iteration, we first compute the group information $\bm{\omega}_t$ and $\bm{\mathcal{G}}_t$ in the lower-level optimization; and then we update the model's parameter in the upper-level optimization. As a result, we obtain the updated parameter $\bm{\theta}_{t+1}$, which is used to compute task-specific risk information at the next iteration. 
In the proposed principle, intra-group tasks share the same scaling weight $\omega^{k} \geq 0$ to prevent similar tasks from inconsistent learning progress, improving knowledge sharing among similar tasks. 

Important in AGRM is to accommodate the grouping during training dynamically. 
In practice, the proposed principle is compatible with any gradient-based optimizer, such as SGD and Adam~\cite{kingma2014adam}, yielding dynamical training for each task. As a new member of the loss-oriented branch, the grouping assignment matrix and group weights in \ours{} can sufficiently utilize the loss information over time to adaptively assign tasks and weight groups.

Unlike prior works on multi-task grouping \cite{fifty2021efficiently, kang2011learning}, which require group-specific architectures, our proposed principle focuses on group-specific scaling and adaptively executing grouping operations. Considering the differences in architecture and optimization within grouping mechanisms, we provide two insights: (i) the criteria for grouping tasks should take both \textit{learning dynamics} and \textit{loss scales} into consideration so that similar tasks can benefit from each other's intermediate feature information and boost performance; (ii) adaptive grouping \textit{aligns learning progress across tasks} and provides a more effective way for across-task information transfer.

\subsection{Dynamical Group Assignment}
\label{sec: dynamic_group_assignment}

In solving the optimization problem in Eq. (\ref{eq: group-ERM}), the main challenge lies in the involvement of discrete and continuous variables, which are implicitly entangled in the objective.
In detail, the lower-level optimization requires adaptively adjusting the discrete variable $\bm{\mathcal{G}}_t$ and the continuous variable $\bm{\omega}_t$ for the learning progress alignment such that the model's parameter $\bm{\theta}$ updates from the lastest grouping information.


Before executing the \textit{lower-level} optimization, we need to introduce task-specific group indicators $\bm{\gamma}_t(\bm{\theta}_t) = \left[\gamma^1_t({\bm{\theta}}_t^s, {\bm{\theta}}_t^1), \gamma^2_t({\bm{\theta}}_t^s, {\bm{\theta}}_t^2), \cdots, \gamma^M_t({\bm{\theta}}_t^s, {\bm{\theta}}_t^M)\right]^{\top} \in \mathbb{R}^{M}$. In general, this involves the entanglement of the model's parameters, which is obtained from high-level optimization.
These group indicators work for exploiting cross-task correlations along the learning progress and provide group information to enable task interactions in the \textit{lower-level} optimization.
We will further discuss the design of the group indicator in Sec.~\ref{sec: calibration}.

Intuitively, we conduct the group assignment as a clustering process based on these group indicators $\bm{\gamma}_t(\bm{\theta}_t)$. 
In this case, each cluster represents a group, and the cluster center is set to the group weight. 
Many clustering algorithms are available to achieve this. In this work, we take the K-means clustering algorithm \cite{krishna1999genetic, kodinariya2013review, ahmed2020k} as a practical clustering implementation. 
Thus, we specify the optimization objective of the dynamical group assignment as:
\begin{equation}
\min_{\bm{\omega}, \bm{\mathcal{G}}} J(\bm{\omega},\bm{\mathcal{G}}; \bm{\theta}_t) := {|| {\bm{\gamma}_t^{\top}}(\bm{\theta}_t) - {\bm{\omega}^{\top} {\bm{\mathcal{G}}} }  ||}^2, 
\label{eq: objective of K-means}
\end{equation}
where $\bm{\omega}_t^{\top} = {\bm{\gamma}_t^{\top}}(\bm{\theta}_t) \bm{\mathcal{G}}_t^{-1}$ indicates that the cluster center closely relates to the group assignment matrix. \textcolor{camera_ready}{It is worth noting that $\bm{\mathcal{G}}_t^{-1}$ is a generalized inverse, especially one-sided right inverse, $\bm{\mathcal{G}}^{-1}=\bm{\mathcal{G}}^{\top}(\bm{\mathcal{G}}\bm{\mathcal{G}}^{\top})^{-1}.$}
The designed dynamical group assignment plays an important role in the lower-level optimization of the proposed AGRM, and it tends to cluster similar tasks into the same group while scattering dissimilar ones in clusters. 

By integrating the dynamical group assignment in Eq. (\ref{eq: objective of K-means}) and the group indicators into  Eq. (\ref{eq: group-ERM}), we provide an instantiation for the AGRM's optimization objective:
\begin{equation}
\begin{aligned}
        \operatorname*{min}_{\bm{\theta}} 
        {{{\bm{\omega}}_t}^{\top}} {\bm{\mathcal{G}}}_t \hat{\bm{L}}(\bm{\theta}) 
         ~\textit{s.t.}~\{{{\bm{\omega}}_t},{\bm{\mathcal{G}}}_t \}  = \operatorname*{arg~min}_{{\bm{\omega}}, \bm{\mathcal{G}}}
        {|| {\bm{\gamma}_t^{\top}}(\bm{\theta}_t) - {\bm{\omega}^{\top} {\bm{\mathcal{G}}} }  ||}^2. \\
\end{aligned}
\label{eq: reformulated-group-ERM}
\end{equation} 

Moreover, the dynamic group assignment heuristically clusters tasks from the group indicators, which avoids the exhausted search of appropriate task combinations for performance gains like previous works ~\cite{fifty2021efficiently}.



\subsection{Risk-guided Group Indicators}
\label{sec: calibration}

This subsection discusses the appropriate design of the group indicators for dynamic group assignment introduced in Sec. \ref{sec: dynamic_group_assignment}. 

The misalignment of learning across tasks can usually be attributed to various risk scales and asynchronous learning dynamics among tasks over time. To address this, we take two operations with risk information, \textit{scale-balance} and \textit{smooth-alignment}, into consideration and then obtain the risk-guided group indicators by combining them. \textcolor{camera_ready}{The role of the group indicators is to use the risk information to explore the relationships among tasks without incurring the expensive computational cost associated with gradients.
Compared with other loss-oriented methods, our group indicator can capture the differences in the per-task risk scale and fully utilize the learning dynamics over time, yielding better representations of risk information. }

\noindent \textbf{Scale-balance.} 
To alleviate the misalignment caused by differences in per-task risk scales, we introduce \textit{scale-balance}, which enlarges the importance of tasks with smaller risks in optimization.
Given task-specific risks at iteration $t$, we normalize them to their average risk for efficient scale balancing. In each iteration, the scale vector for all tasks is denoted as $\mathcal{P}_t(\bm{\theta}_t) =\left[p_{t}^{1}(\bm{\theta}_{t}), p_{t}^{2}(\bm{\theta}_{t}), \dots,p_{t}^{M}(\bm{\theta}_{t})\right]^{\top} \in \mathbb{R}^M 
$, which can be calculated as:
\begin{equation}
\begin{aligned}
    {\mathcal{P}_t(\bm{\theta}_t) ={\rm{diag}}(\hat{\bm{L}}(\bm{\theta}_t))^{-1} \Big[\bar{\hat{\bm{L}}}(\bm{\theta}_t)\Big]_M,}
\end{aligned}
\label{eq: scale_balance_weight}
\end{equation}
where $\rm{diag}(\cdot)$ constructs a diagonal matrix with the elements of the vector placed on the diagonal. $\bar{\hat{\bm{L}}}(\bm{\theta}_t)$ is a scalar to represent the average risk, and $[\cdot]_{M}$ represents the construction of an $M$-dimentional vector whose elements are all equal to the average risk. 
To avoid the upper-level optimization degenerating into a fixed scalar, the gradients of empirical risks in the lower-level optimization are not being computed through them.
However, in practice, the learning dynamics over time tend to make the scale vector inconsistent over iterations~\cite{liu2021conflict, yu2020gradient}, which is not conducive to aligning learning progress. We therefore introduce \textit{smooth-alignment} to update the scale vector with historical information from previous iterations.

\noindent \textbf{Smooth-alignment}.
To avoid sudden fluctuations of scale vectors over iterations, 
we introduce the smoothness vector $\mathcal{Q}_t(\bm{\theta}_{1:t})=\left[q_{t}^{1}(\bm{\theta}_{1:t}), q_{t}^{2}(\bm{\theta}_{1:t}), \dots,q_{t}^{M}(\bm{\theta}_{1:t})\right]^{\top} \in \mathbb{R}^M$, which smooths the updating of the scale vector with previous risk information. Thus, the smoothness vector can deal with asynchronous learning dynamics over time, which helps the model reduce the imbalance of training across tasks. 
To be specific, we compute the smoothness vector by a normalized exponential moving average as follows:
\begin{equation}
\begin{aligned}
\mathcal{Q}_t(\bm{\theta}_{1:t}) &= \sigma \Big[{\mathcal{Q}}_{t-1}(\bm{\theta}_{1:t-1}) \odot \exp(-\beta \hat{\bm{L}}(\bm{\theta}_t))\Big],
\end{aligned}
\label{eq: smooth_alignment_weight}
\end{equation}
where $\odot$ denotes the element-wise multiplication and $\sigma [\cdot]$ normalizes the sum of all smoothness elements to be $1$.
$\beta$ is a temperature hyperparameter to control the influence of current risk information.
Note that when $\beta$ is close to zero, each element in the smoothness vector will degrade to a fixed value $\frac{1}{M}$, which does not capture any historical information to group indicators.

\noindent \textbf{Risk-guided Group Indicators}.
By element-wise multiplying the scale vector in Eq.~(\ref{eq: scale_balance_weight}) and the smoothness vector in Eq.~(\ref{eq: smooth_alignment_weight}), we obtain the group indicators with sufficient risk information as:
\begin{equation}
\begin{aligned}
 & \bm{\gamma}_t(\bm{\theta}_t) = {\mathcal{P}_t(\bm{\theta}_t) \odot \mathcal{Q}_t(\bm{\theta}_{1:t})}.
\end{aligned}
\label{eq: group indicator}
\end{equation}
Then, with the group indicators $\bm{\gamma}_t(\bm{\theta}_t)$,  we optimize the dynamical group assignment in Eq.~(\ref{eq: objective of K-means}) to assign tasks into groups in the lower-level optimization. 

For each group indicator, the role of $\mathcal{P}_t(\bm{\theta}_t)$ in Eq.~(\ref{eq: scale_balance_weight}) and $\mathcal{Q}_t(\bm{\theta}_{1:t})$ in Eq.~(\ref{eq: smooth_alignment_weight}) differs in optimization: the smoothness vector requires the accumulated loss information from previous iterations, while the scale vector are independent of iterations. 
Thus, the smoothness vector can iteratively exploit more consistent task correlations to better align learning progress across tasks.
The experimental section will show that the risk-guided group indicators empirically boost the proposed adaptive group risk minimization in aligning learning tasks.

%% file: fig/demo2.tex
\begin{wrapfigure}{r}{0.5\textwidth}
\centering
\vspace{-14pt}
\includegraphics[width=0.48\textwidth]{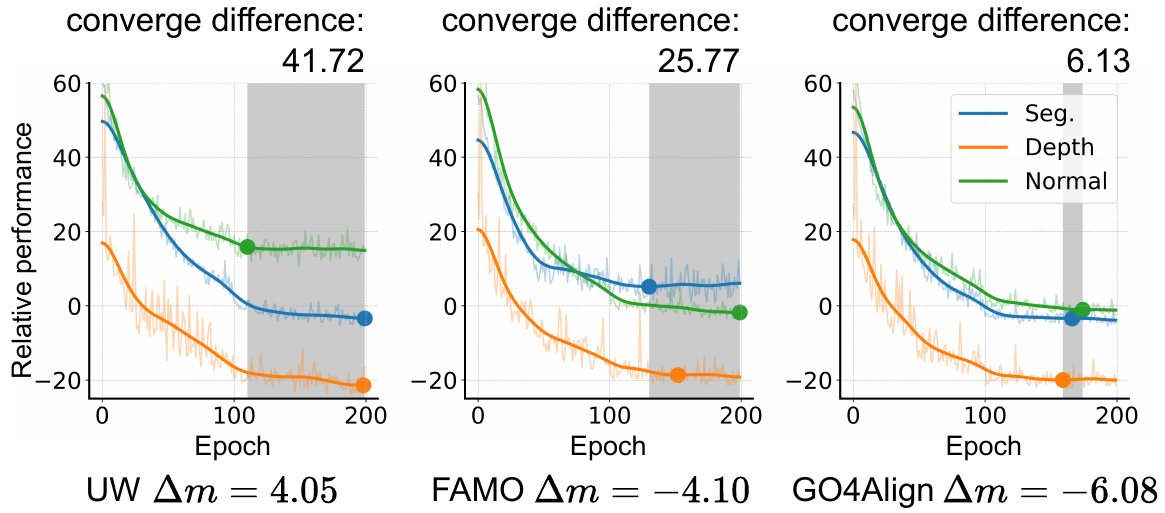}
\vspace{-5pt}
\caption{\textbf{Multi-task alignment and effects on performance.} We visualize relative task performance curves (lower is better) over training epochs. 
Better overall performance usually occurs with lower convergence differences. Our method effectively reduces the convergence difference and achieves a better overall performance.
} 
\label{fig: fig2}
\vspace{-12pt}
\end{wrapfigure}

%% file: fig/GO4Align.tex
\begin{figure*}[t]
\centering
\vspace{-5pt}
\includegraphics[width=1.0\linewidth]{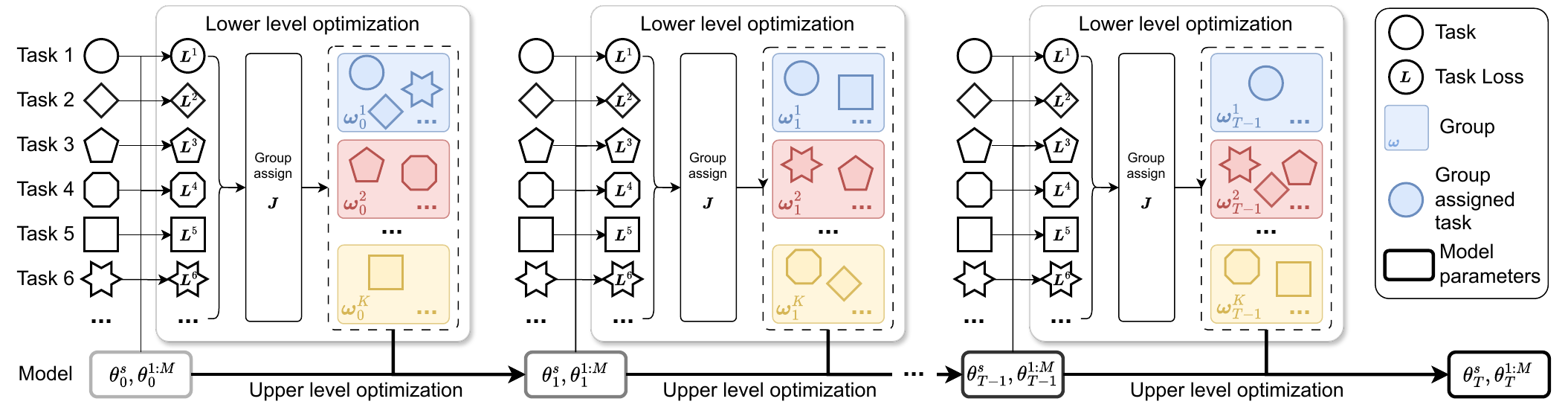}
\vspace{-18pt}
\caption{\textbf{\ours{} using adaptive group risk minimization in the bi-level optimization framework}. In the lower-level optimization, the model assigns tasks to groups with different group weights, encouraging task interactions and aligning learning progress. Such group information is nested into the upper-level optimization for updating the multi-task model's parameters.}
\label{fig: graphmodel}
\vspace{-16pt}
\end{figure*}

%% file: section/3_relatedworks.tex
\section{Related Work}

\noindent\textbf{Multi-Task Optimization.} 
Multi-task optimization addresses the task imbalance issue in multi-task learning, where each task usually influences a shared network differently. \textcolor{camera_ready}{
Task-imbalance in MTO ~\cite{vandenhende2021multi, chen2018gradnorm} refers to imbalanced optimization rather than uneven data distributions in the task space.
}
According to different manipulations in optimization, we roughly divide MTO methods into two branches: 
(i) \textit{gradient-oriented} methods, which solve the task balancing problem by fully utilizing the gradient information of the shared network from different tasks. 
Some studies report impressive performance based on Pareto optimal solutions~\cite{sener2018multi}, gradient normalization~\cite{chen2018gradnorm}, gradient conflicting~\cite{yu2020gradient}, gradient sign Dropout~\cite{chen2020just}, conflict-averse gradient~\cite{liu2021conflict}, Nash bargaining solution~\cite{navon2022multi}. 
However, most gradient manipulation methods usually suffer from high computational cost~\cite{kurin2022defense}.
(ii) \textit{loss-oriented} methods, which reweight task-specific losses with the help of inductive biases from the loss space, e.g., using homoscedastic uncertainty~\cite{kendall2018multi}, task prioritization~\cite{guo2018dynamic}, self-paced learning~\cite{murugesan2017self}, similar learning paces~\cite{liu2019end, liu2023famo}, random loss weight~\cite{lin2021reasonable}. 
Although loss-oriented methods are more computationally efficient, they often underperform gradient-oriented ones in most multi-task benchmarks. 
This paper tries to trade off the overall performance and computational efficiency. 

Recent work~\cite{nguyen2023task} weights tasks under the meta-learning setup but has lower training-time efficiency for large-scale systems with high dimensional parameter space, such as deep neural networks, limiting their applications for dense prediction tasks in MTL.
The closest method to ours is the recent work FAMO~\cite {liu2023famo}, which balances task-specific losses by decreasing task loss approximately at an equal rate. However, \ours{} proposes a new MTL optimizer that dynamically aligns learning progress across tasks by introducing group-based task interactions.

\noindent\textbf{Multi-Task Grouping.} 
Multi-task grouping~\cite{standley2020tasks, song2022efficient, fifty2021efficiently} assigns tasks into different groups and trains intra-group tasks together in a shared multi-task network. 
Previous work~\cite{standley2020tasks} first evaluates the transferring gains for $2^M-1$ candidate multi-task networks ($M$ is the number of tasks) and then conducts the brute-force search for the best grouping.
Some works follow high-order approximation (HOA)~\cite{standley2020tasks} to reduce the prohibitive computational cost. However, they also suffer from inaccurate estimations due to non-linear relationships between high-order gains and corresponding pairwise gains~\cite{song2022efficient}. Meanwhile, \citet{yao2019robust} represents a clustered multi-task learning method, which clusters tasks into several groups by learning the representative tasks. 
The benefit of multi-task grouping is performance gains by training similar tasks together, and this inspires us to capture helpful group information in multi-task optimization. Also, rather than employing different multi-task networks, \ours{} introduce group-based task interactions in scaling for multi-task alignment. 
Moreover, we share a high-level idea of task clustering with ~\cite{thrun1998clustering}. However, task clustering in ~\cite{thrun1998clustering} is limited to pairwise relationships among tasks; meanwhile, our work allows grouping-based task interactions, thus capturing more complex relationships among tasks.


%% file: section/4_experiments.tex
\section{Experiments}
\label{sec: experiment}

\subsection{Comparisons on MTL Benchmarks}
\label{sec: mtsl}

\noindent \textbf{Datasets and Settings.}
We conduct experiments on four benchmarks commonly used in multi-task optimization literature~\cite {liu2023famo,liu2021conflict,liu2019end,navon2022multi}: \texttt{NYUv2}~\cite{SilbermanECCV12}, \texttt{CityScapes}~\cite{cordts2016cityscapes}, \texttt{QM9}~\cite{blum}, and \texttt{CelebA}~\cite{liu2015faceattributes}. For all benchmarks, we follow the training and evaluation protocol in \cite{navon2022multi, liu2023famo}.

\noindent \textbf{Baselines.} We compare \ours{} with a single-task learning baseline, six gradient-oriented methods, and six loss-oriented methods. 
Note that single-task learning (\stl{}) trains an independent deep network for each task.
The gradient-oriented methods include \mgda{}~\cite{sener2018multi}, \pcgrad{}~\cite{yu2020gradient}, \cagrad{}~\cite{liu2021conflict}, \imtlg{}~\cite{liu2020towards}, \graddrop{}~\cite{chen2020just}, and \nashmtl{}~\cite{navon2022multi}.
As for the loss-oriented methods, they are Linear scalarization (\ls{}), Scale-invariant (\si{}), Dynamic Weight Average (\dwa{})~\cite{liu2019end}, Uncertainty Weighting (\uw{})~\cite{kendall2018multi}, Random Loss Weighting (\rlw{})~\cite{lin2021closer}, and \famo{}~\cite{liu2023famo}.
Detailed information about datasets and baselines is in Appendix~\ref{sec: set_up}.

\input{table/nyu}

\input{fig/timecost}

\noindent \textbf{Evaluations.} 
Following previous work \citep{navon2022multi, maninis2019attentive, liu2021conflict}, we report two MTL metrics that demonstrate the overall performance over various task-specific metrics:
(1) \dm{} is the average per-task performance drop relative to STL. We assume there are $S$ metrics for all tasks. $\mathcal{M}^s$ denote the $s$-th metric value of a multi-task method, while $\mathcal{B}^s$ is the corresponding metric value of the STL baseline. Thus, we formulate the average relative performance drop as:${\bm{\Delta m}} \%= \frac{1}{S} \sum_{s=1}^{S} (-1)^{\delta^s} \frac{(\mathcal{M}^s - \mathcal{B}^s)}{\mathcal{B}^s}$,
where ${\delta^s} =1$ if higher values for the $s$-th metric are better and $0$ otherwise. 
(2) $\mr{} = \frac{1}{S} \sum_{s=1}^{S} \text{rank}(\mathcal{M}^s)$ is the average rank of all task-specific metrics, where $\text{rank}(\mathcal{M}^s)$ denotes the ranking of the $s$-th metric value of the model among all comparison methods.
Note that in practice \textit{the lower \dm{} and {\bf{\mr{}}},  the better overall performance}.  

\input{table/qm9}

\noindent \textbf{Effectiveness Comparison.}
We provide performance comparisons on \texttt{NYUv2} in Table~\ref{tab: nyu-v2}. In this benchmark, our method achieves the best overall MTL performance among both gradient-oriented and loss-oriented methods. We observe that our work is the only one that improves each task's performance relative to the corresponding STL performance. This suggests that grouping-based task interactions can adequately alleviate the imbalance of learning progress across tasks.

The experimental results on \texttt{QM9}, \texttt{CityScapes} and \texttt{CelebA} are reported in Table~\ref{tab: qm9-cityscapes-celeba}.  \ours{} obtains the lowest \dm{} on \texttt{QM9}. It also shows comparable performance with \famo{} on \texttt{CityScapes}, one possible reason could be that this dataset only contains $2$ tasks, which limits the potential of the grouping mechanism in our method. In \texttt{CelebA}, even though our work does not achieve the lowest average performance drop, it outperforms all loss-oriented methods, which further verifies the effectiveness of the proposed method. 

\noindent \textbf{Efficiency Comparison.} To show the computational efficiency, in Fig.~\ref{fig: efficiency comparisons}, we report the average training time per epoch over 5 epochs for each method. 
\textcolor{camera_ready}{
We choose LS as the relative baseline for training time (cf. RLW~\cite{lin2021closer} and FAMO~\cite{liu2023famo}) as it is a commonly used MTL baseline with equal weights for each task, and it does not require additional loss-oriented or gradient-oriented techniques.
}
We note that we run all experiments on an NVIDIA A100 and the code of baseline methods comes from ~\citet{liu2023famo} and ~\citet{navon2022multi}. 

As shown in this figure, the proposed method \ours{}, as a new member of the loss-oriented branch, can perform more efficiently than most gradient-oriented methods. \textcolor{camera_ready}{Moreover, when the number of tasks scales up from $2$ to $40$,
the reduction in computational cost between our method and other gradient-oriented methods becomes increasingly significant, e.g., NashMTL~($2.07$) versus Ours~($1.01$) with $2$ tasks, NashMTL~($12.49$) versus Ours~($1.01$) with $40$ tasks.} The main reason is that the training time of gradient-oriented methods is proportional to the number of tasks, but our work can avoid this. More experimental results are provided in Appendix~\ref{sec: additional experimental results}.

\subsection{Ablation Study}
The effectiveness and efficiency of our proposed method are shown in Sec.~\ref{sec: mtsl}. Next, we answer the following questions with our ablation study: 
(1) Can we quantify the contributions of each phrase?
(2) Can we disentangle the roles of the group assignment and group weights?
(3) Can the proposed AGRM principle seamlessly integrate with existing MTO methods?
(4) Are there practical ways to appropriately configure hyperparameters, \textit{e.g.}, group number $K$?
(5) Why do we choose the K-means clustering in the proposed method?

\input{fig/influence}

\input{table/ablationstudy}

\noindent \textbf{Contributions of Each Phase.} To quantify the contributions of each phase in achieving the proposed AGRM principle on \texttt{NYUv2}, we report the detailed performance of our method in each phase. 
\textcolor{camera_ready}{As the grouping is performed by Eq.(4), the first two rows in Table~\ref{tab: ablation_phase} are the variants of our method without grouping, and the last row is our method with grouping. Table~\ref{tab: ablation_phase} empirically examines the performance gains of the variant with task grouping over without grouping.}

In detail, compared with the scale vector in Eq.~(\ref{eq: scale_balance_weight}), the smoothness vector in Eq.~(\ref{eq: smooth_alignment_weight}) can compromise the performance of the ``Normal'' and ``Depth'' tasks, however, scarifying that of ``Seg.''. 
Based on the scale and smoothness vectors, the proposed method employs dynamical group assignment in Eq.~(\ref{eq: reformulated-group-ERM}) to exploit the grouping-based task interactions, thus well aligning the learning progress of similar tasks ``Depth'' and ``Seg.''. We also observe that our method with both phases can improve the task-specific performance relative to STL. This demonstrates that each phase in the method complements each other, resulting in more balanced performance across tasks.

\noindent \textbf{Influence of Group Assignment Matrix.} To explore the influence of the group assignment matrix $\bm{\mathcal{G}}$, we assume the group number is $2$ on \texttt{NYUv2} and make comparisons with several variants, which have various group assignments with fixed group weights $\bm{\omega}=[0.8, 0.1]$. 
As shown in Fig.~\ref{fig: role_of_group} (a-c), grouping ``Seg." and ``Depth" outperforms other options. The main reason could be that these two tasks are very similar and far away from the ``Normal'' task~\cite{fifty2021efficiently}. We observe that variant (d) with a random grouping strategy shows lower performance than the fixed grouping options (a) and (c), which further implies the importance of appropriate group assignment.
It is worth mentioning that the proposed method in (e) without the prior information of the appropriate group assignment also captures such task correlations and each task can get performance gains compared with STL. This demonstrates that group assignment plays an important role in exploring task correlations over time in the proposed method.

\noindent \textbf{Influence of Group Weights.} To study the influence of group weights, we conduct another visualization in Fig.~\ref{fig: role_of_group} (f-i), where we focus on various group weights $\bm{\omega}$ with the ``optimal" group assignment matrix $\bm{\mathcal{G}}=\{\text{Normal}; \text{Seg}, \text{Depth}\}$. 
We observe that with the fixed group assignment matrix, group weights have effects on the extent of compromising among different groups. On the $\texttt{NYUv2}$ dataset, lower weights for the first groups obtain better overall performance. The variant method (i) with random group weights achieves surprising performance, $1.75\%$, in terms of the average relative performance drop. 
(e) shows that our method also tends to dynamically weight the first group with a high value. This demonstrates that group weights are necessary to align the learning progress of different groups over time.

\input{table/effectivenessofAGRM}

\noindent \textbf{Effect of Adaptive Group Risk Minimization Principle.} 
Empirically, the proposed adaptive group risk minimization principle (AGRM) in Sec.~\ref{sec: AGRM} can be seamlessly integrated with existing MTO methods, taking their updated task weights as group indicators. As detailed in Table~\ref{tab: AGRM+related_works}, MTO methods combined with AGRM consistently show improved performance. \mgda{} with adaptive group risk minimization achieves the biggest improvement gap. Moreover, our method still outperforms others. The reason could be that the designed risk-guided group indicators are more suitable for AGRM by balancing risk scales and exploiting historical information from previous iterations.

\input{fig/group}
\noindent \textbf{Influence of Group Number.} In the proposed method, group number $K$ is an important hyper-parameter, especially when we instantiate the clustering process in dynamical group assignment with $K$-means. In this case, there are many different techniques for choosing the right $K$. To be visualizable, here we apply the conventional elbow method. As shown in Fig.~\ref{fig: groups}, the overall performance (lower is better) of our method in (a) and (b) drops at $2$ and $5$, respectively, after that both reach a plateau when the group numbers increase. Thus, in this paper we set $K=2$ and $K=5$ for \texttt{NYUv2} and \texttt{QM9}.

\textcolor{camera_ready}{
\noindent \textbf{Effect of different clustering methods.} In our main experiments, we employed standard K-means for instantiation. K-means is a widely used clustering approach. To investigate the effect of different clustering methods, we evaluate the impact of using alternative clustering algorithms.}

\textcolor{camera_ready}{
Specifically, we tested our proposed method on \texttt{NYUv2} by substituting K-means with SDP-based clustering~\cite{tepper2017surprising} and spectral clustering~\cite{damle2019simple}. As demonstrated in Table~\ref{tab: clustering_method}, these alternative clustering methods also outperform state-of-the-art approaches (FAMO, $-4.10\%$), particularly by enhancing the performance of each task over STL. Interestingly, our experiments show that the K-means clustering algorithm outperforms spectral and SDP-based clustering methods. }

\input{table/clustering_methods}

%% file: table/nyu.tex
\begin{table*}[t]
    \centering
    \caption{\textbf{Results on \texttt{NYUv2} (3 tasks).} The upper and lower tables categorize baseline methods into gradient-oriented and loss-oriented types, respectively.
    Each experiment is repeated over 3 random seeds, and the mean is reported. The best average result is marked in \textcolor{jiayiblue}{\textbf{bold}}. \mr{} and \dm{} are the main metrics for overall MTL performance. Metrics with $\downarrow$ denote that the lower the better.}
    \resizebox{\textwidth}{!}{%
    \begin{tabular}{lrrrrrrrrrrr}
    \toprule
      &  \multicolumn{2}{c}{Segmentation} & \multicolumn{2}{c}{Depth} & \multicolumn{5}{c}{Surface Normal} & &\\
    \cmidrule(lr){2-3}\cmidrule(lr){4-5}\cmidrule(lr){6-10}
    \textbf{Method} &  \multirow{2}{*}{mIoU $\uparrow$} & \multirow{2}{*}{Pix Acc $\uparrow$} & \multirow{2}{*}{Abs Err $\downarrow$} & \multirow{2}{*}{Rel Err $\downarrow$} & \multicolumn{2}{c}{Angle Dist $\downarrow$} & \multicolumn{3}{c}{Within $t^\circ$ $\uparrow$}  & \mr{} $\downarrow$ &  \dm{} $\downarrow$ \\
    \cmidrule(lr){6-7}\cmidrule(lr){8-10}
    & & & & & Mean & Median & 11.25 & 22.5 & 30  &\\
    \midrule 
    \stl{}       & 38.30 & 63.76 & 0.6754 & 0.2780 & 25.01 & 19.21 & 30.14 & 57.20 & 69.15   & - & -     \\
    \midrule
    \mgda{}      & 30.47 & 59.90 & 0.6070 & 0.2555 & 24.88 & 19.45 & 29.18 & 56.88 & 69.36   & 7.00 & 1.38  \\
    \pcgrad{}    & 38.06 & 64.64 & 0.5550 & 0.2325 & 27.41 & 22.80 & 23.86 & 49.83 & 63.14   & 9.00 & 3.97  \\
    \graddrop{}  & 39.39 & 65.12 & 0.5455 & 0.2279 & 27.48 & 22.96 & 23.38 & 49.44 & 62.87   & 7.89 & 3.58  \\
    \cagrad{}    & 39.79 & 65.49 & 0.5486 & 0.2250 & 26.31 & 21.58 & 25.61 & 52.36 & 65.58   & 5.33 & 0.20  \\
    \imtlg{}     & 39.35 & 65.60 & 0.5426 & 0.2256 & 26.02 & 21.19 & 26.20 & 53.13 & 66.24   & 4.56 & -0.76  \\
    \nashmtl{}   & 40.13 & \best{65.93} & \best{0.5261} & 0.2171 & 25.26 & 20.08 & 28.40 & 55.47 & 68.15   & 2.89 & -4.04  \\
    \midrule
    \ls{}        & 39.29 & 65.33 & 0.5493 & 0.2263 & 28.15 & 23.96 & 22.09 & 47.50 & 61.08   & 9.89 & 5.59  \\
    \si{}        & 38.45 & 64.27 & 0.5354 & 0.2201 & 27.60 & 23.37 & 22.53 & 48.57 & 62.32   & 8.78 & 4.39  \\
    \rlw{}       & 37.17 & 63.77 & 0.5759 & 0.2410 & 28.27 & 24.18 & 22.26 & 47.05 & 60.62   &12.22 & 7.78  \\
    \dwa{}       & 39.11 & 65.31 & 0.5510 & 0.2285 & 27.61 & 23.18 & 24.17 & 50.18 & 62.39   & 8.67 & 3.57  \\
    \uw{}        & 36.87 & 63.17 & 0.5446 & 0.2260 & 27.04 & 22.61 & 23.54 & 49.05 & 63.65   & 8.33 & 4.05  \\
    \famo{}      & 38.88 & 64.90 & 0.5474 & 0.2194 & 25.06 & 19.57 & 29.21 & 56.61 & 68.98   & 4.33 & -4.10 \\
    \ours{} &\best{40.42}	&65.37	&0.5492	&\best{0.2167}	&\best{24.76}	&\best{18.94}	&\best{30.54}	&\best{57.87}	&\best{69.84}	&\best{2.11}	&\best{-6.08} \\
    \bottomrule  
    \end{tabular}
    }
    \label{tab: nyu-v2}
\end{table*}

%% file: fig/timecost.tex
\begin{figure*}[t]
\centering
\includegraphics[width=1\linewidth]{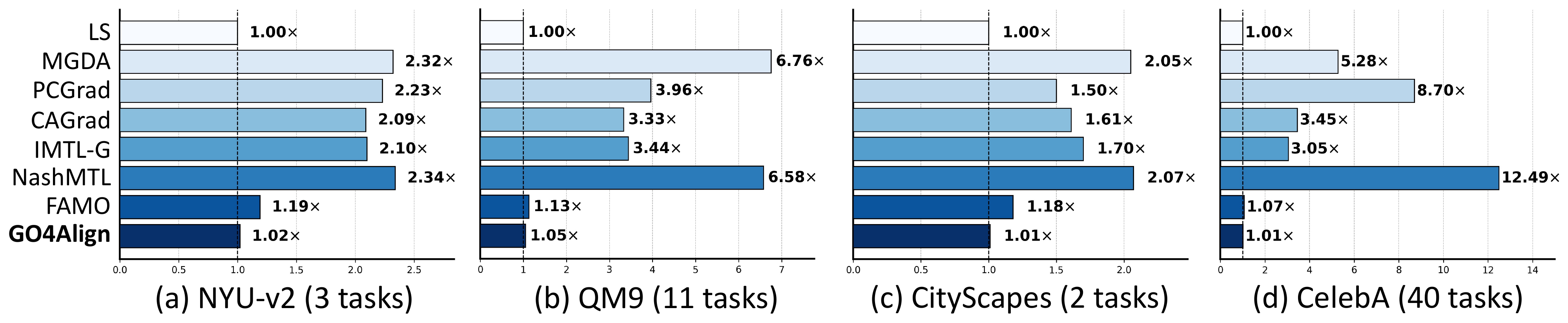}
\caption{\textbf{Efficiency comparisons on training time.} Each method’s training time is relative to a simple baseline method with Eq.~(\ref{eq: scale-ERM}), which minimizes the sum of
task-specific empirical risks.}
\label{fig: efficiency comparisons}
\end{figure*}

%% file: table/qm9.tex
\begin{wraptable}{r}{0.5\textwidth}
    \centering
    \caption{\textbf{Comparisons on \texttt{QM9} (11 tasks), \texttt{CityScapes} (2 tasks) and \texttt{CelebA} (40 tasks).} 
    Detailed results are in Appendix \ref{sec: additional experimental results}.}
    \label{tab: qm9-cityscapes-celeba}
    \resizebox{0.45\textwidth}{!}{%
    \begin{tabular}{lrrrrrr}
    \toprule
   \multirow{2}{*}{\textbf{Method}} & \multicolumn{2}{c}{\textbf{QM9}} & \multicolumn{2}{c}{\textbf{CityScapes}} & \multicolumn{2}{c}{\textbf{CelebA}} \\
      \cmidrule(lr){2-3} \cmidrule(lr){4-5} \cmidrule(lr){6-7}
      & {\mr{} $\downarrow$} & { \dm{} $\downarrow$} & {\mr{} $\downarrow$} & { \dm{} $\downarrow$}  & {\mr{} $\downarrow$} & { \dm{} $\downarrow$} \\
    \midrule
    \mgda{}     &7.73 & 120.5  & 10.00 & 44.14  & 10.05 & 14.85\\
    \pcgrad{}   &6.09 & 125.7 & 6.25 & 18.29  & 6.05 & 3.17 \\
    \cagrad{}   & 7.09 & 112.8  & 5.00 & 11.64  & 5.65 & 2.48 \\
    \imtlg{}    & 5.91 &  77.2 & 4.00 & 11.10  & 4.08 & \best{0.84}\\
    \nashmtl{}  &\best{3.64} &  62.0 & \best{2.50} & 6.82  & 4.53 & 2.84 \\
    \midrule
    \ls{}       & 8.00 & 177.6 & 8.50 & 14.11 & 5.55 & 4.15  \\
    \si{}       & 5.09 &  77.8 & 8.50 & 14.11 & 7.10 & 7.20  \\
    \rlw{}      & 9.36 & 203.8 & 7.75 & 24.38 & 4.60 & 1.46  \\
    \dwa{}      & 7.64 & 175.3 & 6.00 & 21.45 & 6.25 & 3.20  \\
    \uw{}       & 6.64 & 108.0 & 5.75 & \best{5.89} & 5.18 & 3.23 \\
    \famo{}     & 4.73 & 58.5 & 5.50  & 8.13 & 4.10 & 1.21  \\
    \ours{}     &4.55 &\best{52.7}&7.00 &8.11 &\best{3.10} &0.88\\
    \bottomrule 
    \end{tabular}
    }
\end{wraptable}

%% file: fig/influence.tex
\begin{figure}[t]
\centering
\includegraphics[width=1.0\linewidth]{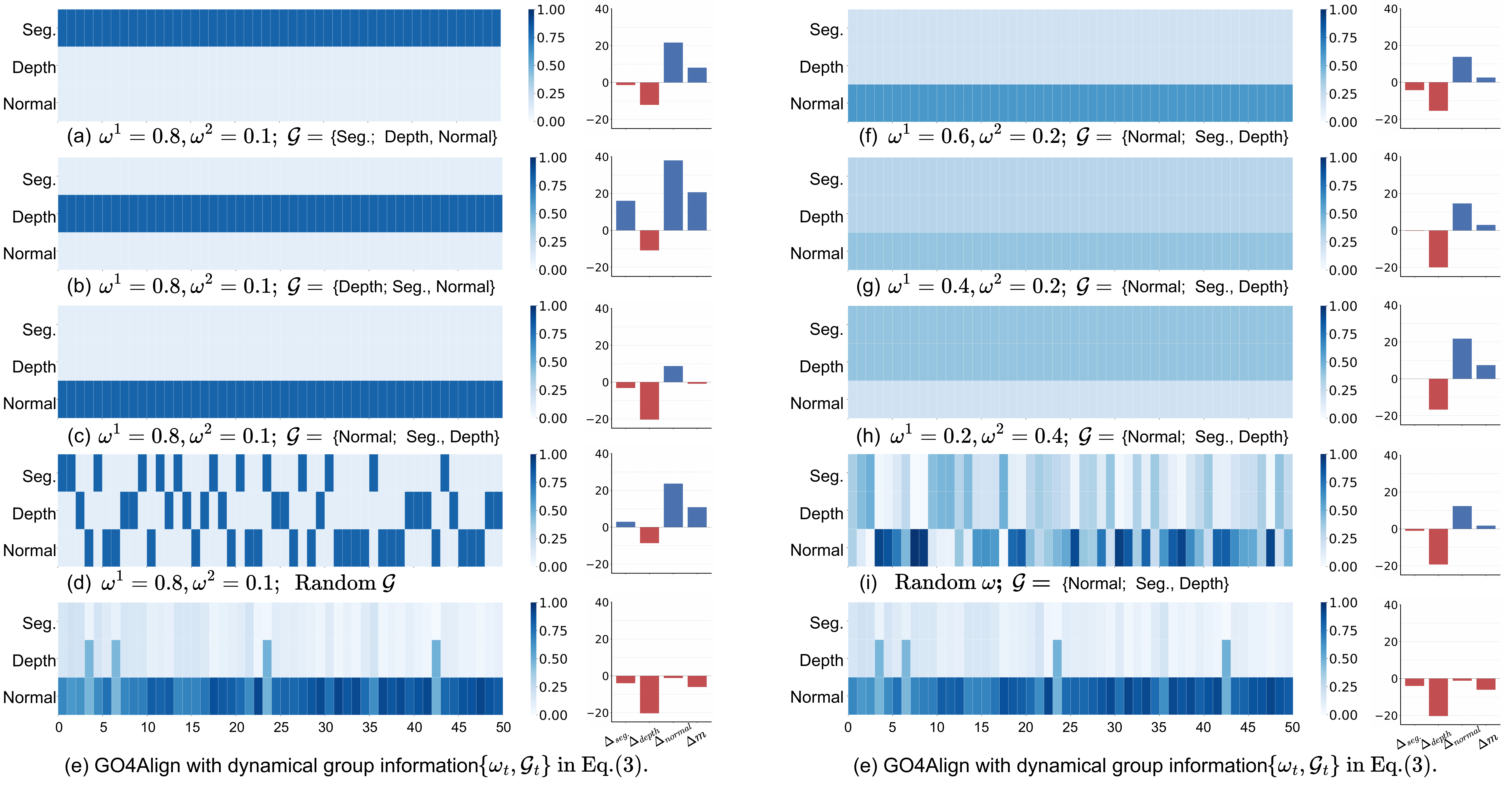}
\vspace{-25pt}
\caption{\textbf{Comparative analysis of the influence of the group assignment matrix and group weights on} \texttt{NYUv2}. 
\textcolor{camera_ready}{The x-axis in the subplots denotes the epoch, and the intensity of the color indicates the weight value.}
\textbf{(a-d)} have fixed group weights $\bm{\omega}=[\omega^1, \omega^2]$ but various group assignment matrices $\mathcal{G}$. 
\textbf{(f-i)} have various group weights $\bm{\omega}$ but a fixed group assignment matrix $\mathcal{G}$. 
\textbf{(e)} is our method that dynamically exploits a group assignment matrix and group weights for each iteration. The right side of each method shows relative performance drops on each task and their average one.}
\label{fig: role_of_group}
\vspace{-15pt}
\end{figure}

%% file: table/ablationstudy.tex
\begin{wraptable}{r}{0.5\textwidth}
\centering
\vspace{-10pt}
\caption{\textbf{Effectiveness of each phase in \ours{} on}~\texttt{NYUv2}. \ymark~denote whether the component joins the pipeline.}
\resizebox{1.0\linewidth}{!}{%
\begin{tabular}{ccccccc}
\toprule
Eq.(\ref{eq: scale_balance_weight}) &  Eq.(\ref{eq: smooth_alignment_weight})  & Eq.(\ref{eq: reformulated-group-ERM}) & $\bm{\Delta}_{seg.}\%$ $\downarrow$ & $\bm{\Delta}_{depth}\%$ $\downarrow$ & $\bm{\Delta}_{normal}\%$ $\downarrow$ & \dm{} $\downarrow$ \\
\midrule
\ymark &  & & -0.02	&\best{-21.76}	&13.14 & 2.46	\\
\ymark & \ymark & &14.22 &-15.27 &2.52 & 1.16\\
\ymark & \ymark & \ymark &\best{-4.03} &-20.37 &\best{-1.18} & \best{-6.08}\\
\bottomrule
\end{tabular}}
\label{tab: ablation_phase}
\end{wraptable}

%% file: table/effectivenessofAGRM.tex
\begin{wraptable}{r}{0.5\textwidth}
\centering
\vspace{-6.5mm}
\caption{\textbf{Comparisons of existing MTO methods with the proposed AGRM on}~\texttt{NYUv2}.}
\resizebox{1.0\linewidth}{!}{%
\begin{tabular}{lcccc}
\toprule
\textbf{Methods} & $\bm{\Delta}_{seg.}\%$ $\downarrow$ & $\bm{\Delta}_{depth}\%$ $\downarrow$ & $\bm{\Delta}_{normal}\%$ $\downarrow$ & $\bm{\Delta}_{m}\%$ $\downarrow$\\
\midrule
\mgda{} &13.25	&-9.11	&0.83 & 1.38\\
\mgda{} + AGRM & 6.06  & -11.69 & -1.14 & -1.89\\
\midrule
\nashmtl{} &-4.09	& \best{-22.01}	&3.15 & -4.04\\
\nashmtl{} + AGRM & \best{-7.75}  & -20.14  & 3.60 & -4.20 \\
\midrule
\famo{} &-1.65	&-20.02	&1.29 & -4.10\\
\famo{} + AGRM & 1.76 & -21.17 & -0.03 & -4.32\\
\midrule
\ours{} &{-4.03} &-20.37 
&\best{-1.18}  & \best{-6.08}\\
\bottomrule
\end{tabular}}
\vspace{-10pt}
\label{tab: AGRM+related_works}
\end{wraptable}

%% file: fig/group.tex
\begin{wrapfigure}{r}{0.5\textwidth}
\centering
\vspace{-1pt}
\includegraphics[width=1\linewidth]{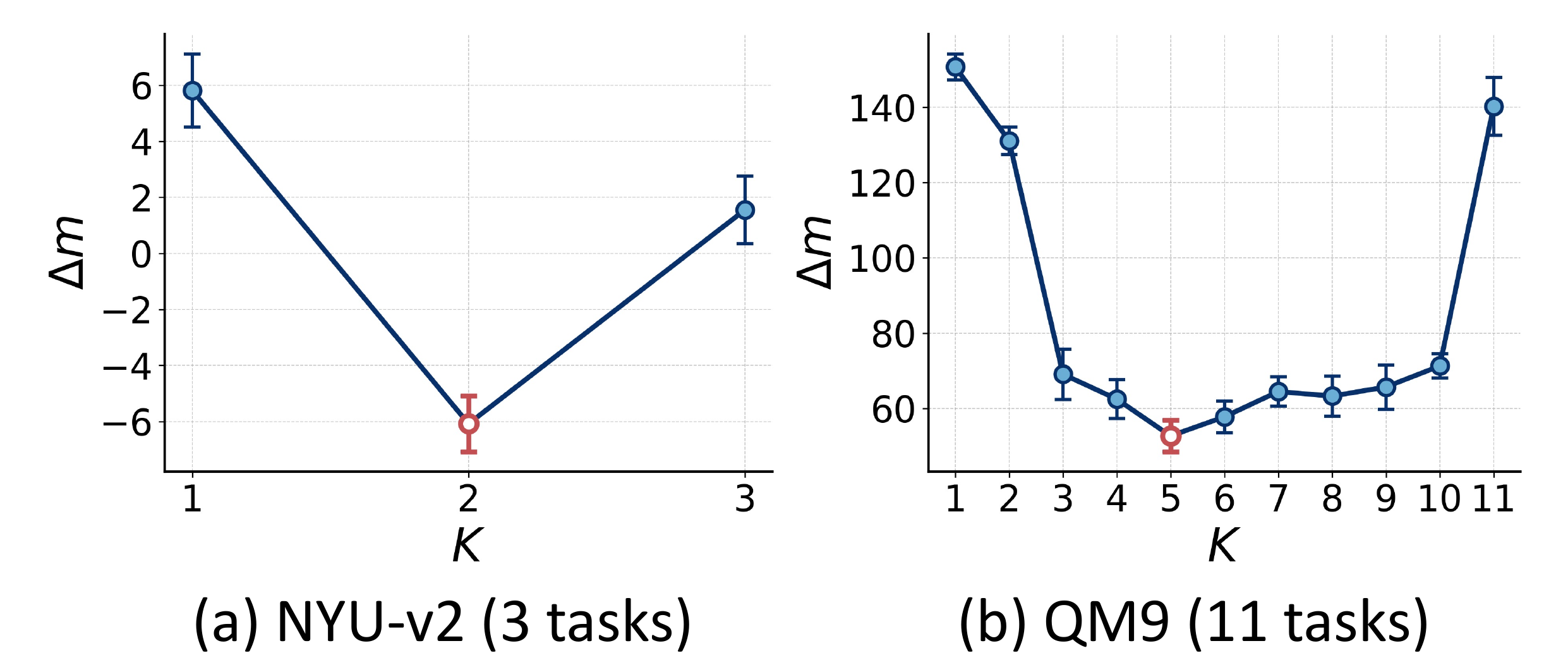}
\vspace{-10pt}
\caption{\textbf{Identification of ``elbow'' points on } \texttt{NYUv2} \textbf{and} \texttt{QM9}. According to the conventional elbow method, we set the group number of the two datasets as 2 and 5, respectively.}
\label{fig: groups}
\vspace{-15pt}
\end{wrapfigure}

%% file: table/clustering_methods.tex
\begin{table}
\centering
\caption{\textbf{Comparisons between different clustering methods in the proposed framework.}}
\resizebox{1.0\textwidth}{!}{%
\begin{tabular}{lccccc}
\toprule
\textbf{Methods} & $\bm{\Delta}_{seg.}\%$ $\downarrow$ & $\bm{\Delta}_{depth}\%$ $\downarrow$ & $\bm{\Delta}_{normal}\%$ $\downarrow$ & $\bm{\Delta}_{m}\%$ $\downarrow$ & \textbf{Relative runtime} $\downarrow$\\
\midrule
SDP-based clustering~\cite{tepper2017surprising}     
& -2.97   & -18.76  & -1.09   & -5.44  & 1.20$\times$  \\
Spectral clustering~\cite{damle2019simple}      
& -1.78   & -18.58  & -0.06   & -4.56  & 1.17$\times$  \\
Ours                                
& \best{-4.03}   & \best{-20.37}  & \best{-1.18}   & \best{-6.08}  & \best{1.02$\times$}  \\     
\bottomrule
\end{tabular}
}
\label{tab: clustering_method}
\end{table}

%% file: section/5_conclusions.tex
\section{Conclusion}
\label{sec: conclusion}

\noindent \textbf{Technical Discussion.} This paper focuses on the task imbalance issue in MTO.  Previous MTO methods suffer from either intensive computations or non-competitive performance. Our proposed \ours{} addresses the issue by aligning learning progress across tasks with the help of the AGRM principle. In problem-solving, we present a tractable optimization pipeline, which incorporates grouping-based task interactions into the loss scaling of MTO.  

\noindent \textbf{Limitation.} 
The main limitation of this work is the heuristic configuration of the group numbers. Although the search space is significantly smaller than some grouping multi-task methods~\cite{song2022efficient}, it still needs maximum $M$ runs to find the best number. 
Some related techniques~\cite{sinaga2020unsupervised} automatically set the group number can be added to avoid this limitation in future work.

\noindent \textbf{Broader Impact.}
This paper is the first to consider task grouping in multi-task optimization with deep multi-task models. We propose a simple and principled way to fasten multi-task optimization with better training-time efficiency, which has many potential societal impacts, especially in dense prediction tasks. \textcolor{camera_ready}{
We provide the code for our method to encourage follow-up work.\footnote{https://github.com/autumn9999/GO4Align}}

\section*{Acknowledgment}
\textcolor{camera_ready}{
This work is financially supported by the Inception Institute of Artificial Intelligence, the University of Amsterdam and the allowance Top consortia for Knowledge and Innovation (TKIs) from the Netherlands Ministry of Economic Affairs and Climate Policy.}

%% file: section/6_appendix.tex
\newpage
\appendix
\onecolumn

\section*{Appendix}

\section{Algorithm of \ours{}}
\label{sec: algorithm}
The pseudo-code of \ours{} is provided in Algorithm~\ref{algorithm}. 
For clarity, we also illustrate the optimization process in Fig.~\ref{fig: computation graph}. 

\input{table/algorithm}

\input{fig/illustration}

\section{Experimental Set-Up \& Implementation Details}
\label{sec: set_up}

\subsection{Benchmark Descriptions}

\texttt{NYUv2}~\cite{SilbermanECCV12} is an indoor scene dataset consisting of 1449 RGBD images and dense per-pixel labeling with 13 classes. The learning objectives include $3$ different dense prediction tasks: image segmentation, depth prediction, and surface normal prediction based on any scene image.

\texttt{CityScapes}~\cite{cordts2016cityscapes} contains 5000 street-view RGBD images with per-pixel annotations. It needs to predict $2$ dense prediction tasks: image segmentation and depth prediction. 

\texttt{QM9}~\cite{blum} is a benchmark for group neural networks to predict $11$ properties of molecules. It consists of $>$130K molecules represented as graphs annotated with node and edge features. We use $110K$ molecules from the QM9 example in PyTorch Geometric~\cite{fey2019fast}, $10K$ molecules for validation, and the rest of $10K$ molecules as a test set. 

\texttt{CelebA}~\cite{liu2015faceattributes} contains 200K face images of 10K different celebrities, and each face image is provided with 40 facial binary attributes. As the protocol provided in previous work~\cite{liu2023famo}, each attribute corresponds to one task. Thus, we consider CelebA as a 40-task MTL problem. 

\textcolor{camera_ready}{This work follows the same experimental setting used in NashMTL~\cite{navon2022multi} and FAMO~\cite{liu2023famo}, including the dataset partition for training, validation, and testing. The benchmark partition is shown in Table~\ref{tab: dataset}. We also note that NYUv2 and Cityscapes do not have validation sets. Following the protocol in ~\cite{navon2022multi,liu2023famo}, we report the test performance averaged over the last ten epochs.}

\input{table/dataset}

\subsection{Compared Multi-task Learning Baselines}

From the gradient manipulation branch, \textbf{(1)} \mgda{}~\cite{sener2018multi} that finds the equal descent direction for each task; \textbf{(2)} \pcgrad{}~\cite{yu2020gradient} proposes to project each task gradient to the normal plan of that of other tasks and combining them together in the end; \textbf{(3)} \cagrad{}~\cite{liu2021conflict} optimizes the average loss while explicitly controls the minimum decrease across tasks; \textbf{(4)} \imtlg{}~\cite{liu2020towards} finds the update direction with equal projections on task gradients; \textbf{(5)} \graddrop{}~\cite{chen2020just} that randomly dropout certain dimensions of the task gradients based on how much they conflict; \textbf{(6)} \nashmtl{}~\cite{navon2022multi} formulates MTL as a bargaining game and finds the solution to the game that benefits all tasks. 

From the loss scaling branch, \textbf{(1)} Linear scalarization (\ls{}) is the sum of empirical risk minimization; \textbf{(2)} Scale-invariant (\si{}) is invariant to any scalar multiplication of task losses; \textbf{(3)} Dynamic Weight Average (\dwa{})~\cite{liu2019end}, a heuristic for adjusting task weights based on rates of loss changes; \textbf{(4)} Uncertainty Weighting (\uw{})~\cite{kendall2018multi} uses task uncertainty as a proxy to adjust task weights; \textbf{(5)} Random Loss Weighting (\rlw{})~\cite{lin2021closer} that samples task weighting whose log-probabilities follow the normal distribution; \textbf{(6)} \famo{}~\cite{liu2023famo} decreases task losses approximately at equal rates. 

\subsection{Neural Architectures \& Training Details}
\label{sec: appendix/trainingdetails}

For \texttt{NYUv2} and \texttt{CityScapes}, we follow the training and evaluation protocol in \cite{navon2022multi}, which adds data augmentations during training for all compared methods. 
We train each method for 200 epochs with an initial learning rate of $1e-4$ and reduce the learning rate to $5e-5$ after 100 epochs. The architecture is Multi-Task Attention Network (MTAN)~\cite{liu2019end} built upon SegNet~\cite{badrinarayanan2017segnet}. Batch sizes for NYUv2 and CityScapes are set as 2 and 8  respectively. To make a fair comparison with previous works~\cite{liu2019end, liu2021conflict, liu2023famo, yu2020gradient}, we report the test performance averaged over the last 10 epochs.

We follow the protocol in ~\citet{navon2022multi} to normalize each task target at the same scale for fairness. 
We train each method for 300 epochs with a batch size of 120 and search for the best learning rate in ${\{1e-3, 5e-4, 1e-4\}}$. 
We take ReduceOnPlateau~\cite{navon2022multi} as the learning-rate scheduler to decrease the lr once the validation overall performance stops improving. The validation set is also used for early stopping.

Following~\cite{liu2023famo}, we use a neural network with five convolutional and two fully connected layers as the shared encoder.
The decoder of each task is implemented by another fully connected layer.
We train the model for 15 epochs with a batch size of 256. We adopt Adam as the optimizer with a fixed learning rate of $1e-3$.
Similar to \texttt{QM9}, we use the validation set for early stopping and hyperparameter selection, such as the number of groups $K$ and the step size of smoothness value $\beta$. We conduct all experiments on a single NVIDIA A100 GPU.

\section{Additional Experimental Results}
\label{sec: additional experimental results}

\subsection{Detailed Results on \texttt{QM9} and \texttt{CityScapes}}
We provide task-specific performance on \texttt{QM9} in Table~{\ref{tab: qm9_detail}}. The proposed \ours{} obtains the best performance in terms of the average performance drop \dm{}. And its average rank \mr{} is lower than all loss-oriented methods, which demonstrates the proposed method can get a more balanced performance for each task.  

As shown in Table~{\ref{tab: cityscapes_detail}}, our method achieves competitive performance on \texttt{CityScapes} with other alternatives except for \nashmtl{} and \uw{}. The main reason can be that there are only two tasks in the datasets, which constrains the effectiveness of the grouping mechanism in our method.

\input{table/details}

\subsection{Training Time Comparisons}
In Fig.~\ref{fig: efficiency comparisons}, we show the training time of MTO methods relative to a baseline method (LS). Here we provide real training time (seconds) in Table~\ref{tab: trainingtime}, where we compute the average training time of 5 epochs. From this table, we observe that loss-oriented methods in general use less time for one epoch than gradient-oriented methods. \ours{} requires the second lowest time cost during training on all four datasets, demonstrating its good computational efficiency.

\input{table/time}

\subsection{Visualizations on Risk Ratios}
To investigate the influence of different scaling methods on the training of tasks, we illustrate the ratios between task-specific empirical risk and the sum of all empirical risks before and after scaling on \texttt{NYUv2}. In each small figure,  the three shadows from top to down represent the ratios of ``Normal'',  ``Seg.'' and ``Depth'', respectively.
The x-axis represents epochs ranging from $1$ to $200$. Note that \mgda{} and \nashmtl{} scale task-specific gradients. For direct comparisons, here we provide the scaled loss ratios, which are equivalent to gradient scaling for the shared network. 

\begin{figure*}[h]
\centering
\includegraphics[width=1\linewidth]{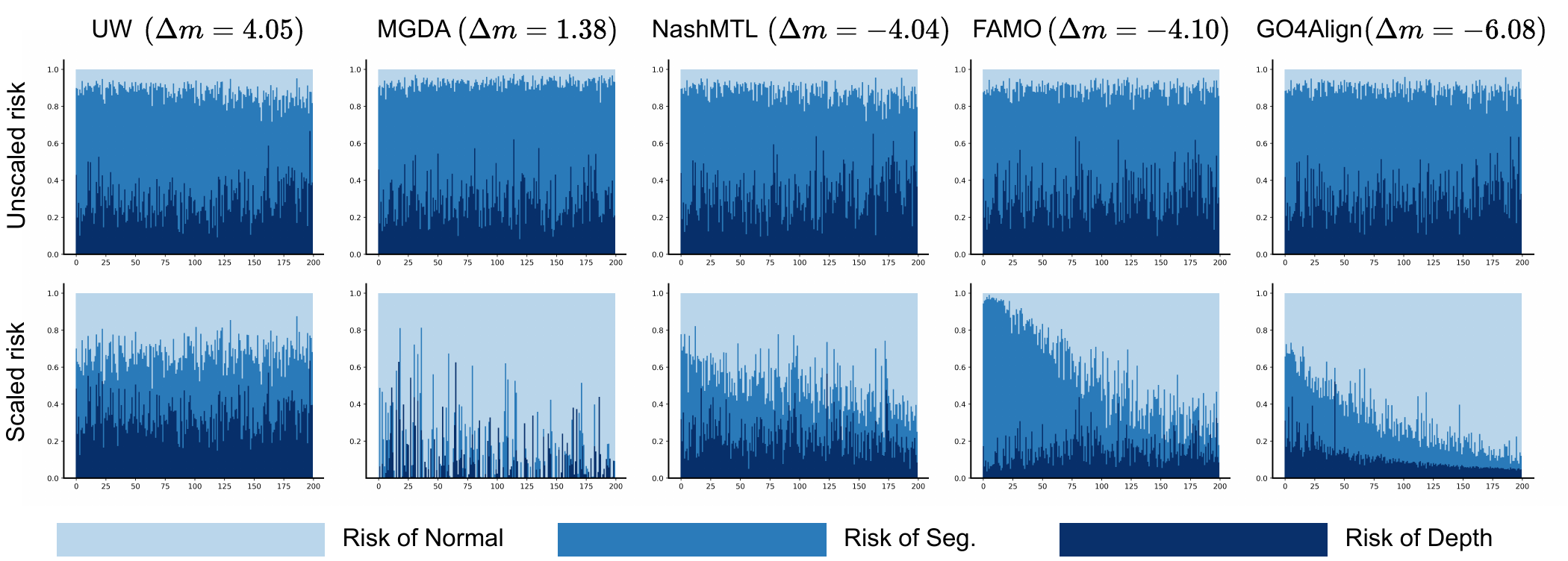}
\caption{\textbf{Analysis on risk ratios.}
Compared with other works, the proposed \ours{} shows more stable ratios among tasks over epochs, indicating that \ours{} can maintain better alignment throughout the training process.}
\label{fig: loss ratio}
\end{figure*}

As shown in Fig.~\ref{fig: loss ratio}, all methods have similar ratios over epochs on the unscaled risks but perform differently on the scaled risks. We observe that \mgda{} have a significantly larger ratio on the ``Normal'' task, which means \mgda{} prefers to optimize the ``Normal'' task. Compared with related works, our method has more stable ratios of different tasks. The possible reason could be that our method benefits from historical information, which avoids training instability among tasks.

\section{Additional Related Work}
In MTL, we group multi-task learning methods into three categories: multi-task optimization, multi-task grouping, and deep multi-task architecture. The first two categories are most related to this paper and are mentioned in the main paper. To be self-contained, here we provide detailed discussions about deep multi-task architecture. 

Our method \ours{} builds on multi-task grouping and multi-task optimization, inheriting advantages from both to balance different tasks in joint learning. In the following, we discuss each category of multi-task learning methods and explain how they relate to our proposed method.

\noindent\textbf{Deep Multi-Task Architecture.} 
Multi-task architecture design can be roughly categorized into either a hard-parameter sharing design~\cite{kokkinos2017ubernet, lu2017fully} or a soft-parameter sharing design~\cite{misra2016cross, gao2019nddr,liu2019end}. 
The hard-parameter sharing design generally contains a shared encoder and several task-specific decoders. 
Branching points between the shared encoder and decoders are determined in an ad-hoc way~\cite{vandenhende2021multi, vandenhende2019branched}, resulting in a suboptimal solution.  
Some work~\cite{guo2020learning, lu2017fully, sun2020adashare} automatically learns where to share or branch with a network. 
The soft-parameter sharing design considers all parameters task-specific and instead learns feature-sharing mechanisms to handle the cross-task interactions~\cite{misra2016cross, long2017learning}.  
Some work~\cite{shen2021variational, shen2023episodic} are probabilistic multi-task learning methods, which introduce latent variables to encode task-specific information and enable knowledge sharing in the latent spaces.
Soft-parameter sharing methods usually struggle with model scalability as the model size grows linearly with the number of tasks. 
\ours{} focuses on the optimization of multi-task learning, which adopts a simple hard-parameter sharing architecture as the backbone and subsequently reduces task interference in the gradient space of the shared encoder or the loss space. 




%% file: table/algorithm.tex
\begin{algorithm}[ht]
    \caption{Group Optimization for Multi-Task Alignment (\ours{})}
    \begin{algorithmic}[1]
        \STATE \textbf{Input}:  
        Maximum iteration number $T$;
        Batch size $N_{\rm{bz}}$;
        Learning rate $\alpha$; 
        Temperature hyperparameter $\beta$;
        Task number $M$;
        Group number $(1 < K \leq M)$;
        
        \STATE Initialize model parameters $\bm{\theta}_0 = \{\bm{\theta}^{s}_0,\bm{\theta}^{1}_0, \bm{\theta}^{2}_0, \cdots, \bm{\theta}^{M}_0\}$;
        

        \FOR{$t = 0 : T$}
            \STATE Randomly sample a batch of training samples: 
            
            $\{ ({\bm{x}}_n, y_n^1, ..., y_n^M)\}_{n=1}^{N_{\rm{bz}}}, \text{where}~ \bm{x}_n \in \mathcal{X}, y_n^m \in \mathcal{Y}^m;$
            
            \STATE Compute the empirical risk for each task at the $t$-th iteration: 
            
            $\hat{\mathcal{L}}^m( \bm{\theta}^{s}_t, \bm{\theta}^{m}_t) = \frac{1}{N_{\rm{bz}}} \sum_{n=1}^{N_{\rm{bz}}} \ell^m(f({\bm{x}}_n; \bm{\theta}^{s}_t, \bm{\theta}^{m}_t), y_n^m);$
            
            \vspace{+2mm}
            {\textcolor{jiayiblue}{\texttt{\textbf{// Lower-level optimization for grouping-based task interactions.}}}}
            
            \STATE Compute the scale vector ${{\mathcal{P}}_t(\bm{\theta}_t)}$ in Eq.~(\ref{eq: scale_balance_weight});

            \STATE Compute the smoothness vector ${{\mathcal{Q}}_t(\bm{\theta}_{1:t})}$ in Eq.~(\ref{eq: smooth_alignment_weight});

            \STATE Obtain the group indicators ${\bm{\gamma}}_t(\bm{\theta}_{1:t})$ with Eq.(\ref{eq: group indicator});
    
            \STATE Update the group information $\bm{\omega}_t$ and $\bm{\mathcal{G}}_t$ with Eq.~(\ref{eq: objective of K-means}) and the obtained ${\bm{\gamma}}_t(\bm{\theta}_{1:t})$.
            
            \vspace{+2mm}
            {\textcolor{jiayiblue}{\texttt{\textbf{// Upper-level optimization for model parameters.}}}}
           \STATE Update the model parameters $\bm{\theta}$ with Eq. (\ref{eq: reformulated-group-ERM}):
            $\bm{\theta}_{t+1} \leftarrow \operatorname*{arg min}_{\bm{\theta}} {{{\bm{\omega}}_t}^{\top}} {\bm{\mathcal{G}}}_t \hat{\bm{L}}(\bm{\theta}).$
        \ENDFOR
    \end{algorithmic}
\label{algorithm}
\end{algorithm}

%% file: fig/illustration.tex
\begin{figure*}[h]
\centering
\includegraphics[width=0.99\linewidth]{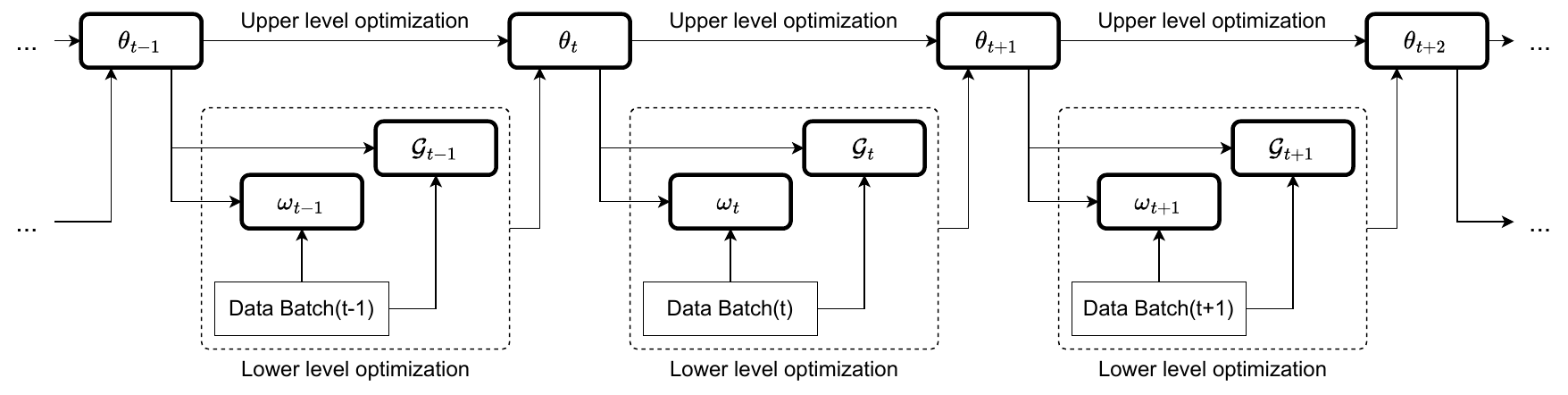}
\caption{\textbf{Optimization process of the proposed adaptive group risk minimization principle.} At each iteration, given the randomly sampled mini-batch data, we first compute the group information $\bm{\omega}$ and $\bm{\mathcal{G}}$ in the lower-level optimization and then update the model's parameter $\bm{\theta}$ in the upper-level optimization.}
\label{fig: computation graph}
\end{figure*}

%% file: table/dataset.tex
\begin{table}[h]
\centering
\caption{\textbf{Benchmark partition for training, validation, and testing.}}
\begin{tabular}{c|c|c|c|c}
\toprule
\multicolumn{1}{c|}{\textbf{Datasets}} & \multicolumn{1}{c|}{\textbf{Total}} & \multicolumn{1}{c|}{\textbf{Training}} & \multicolumn{1}{c|}{\textbf{Validation}} & \multicolumn{1}{c}{\textbf{Test}} \\ 
\midrule
NYUv2                                   & 1449                                & 795                                    & N/A                                      & 654                                \\ 
CityScapes                              & 3475                                & 2975                                   & N/A                                      & 500                                \\ 
QM9                                     & $\sim$130k                          & $\sim$110k                             & 10k                                      & 10k                                \\ 
CelebA                                  & 202,599                             & 162,770                                & 19,867                                   & 19,962                             \\ 
\bottomrule
\end{tabular}
\label{tab: dataset}
\end{table}

%% file: table/details.tex
\begin{table*}[t]
    \centering
    \caption{\textbf{Detailed results on }~\texttt{QM9}. Each experiment is repeated over 3 random seeds and the mean is reported. The best average result is marked in bold. \mr{} and \dm{} are the main metrics for MTL performance.}
    \resizebox{\textwidth}{!}{%
    \begin{tabular}{lrrrrrrrrrrrrr}
    \toprule
    \multirow{2}{*}{\textbf{Method}} & $\mu$ & $\alpha$ & $\epsilon_\text{HOMO}$ & $\epsilon_\text{LUMO}$ & $\langle R^2\rangle$ & ZPVE & $U_0$ & $U$ & $H$ & $G$ & $c_v$ &\multirow{2}{*}{\mr{} $\downarrow$} & \multirow{2}{*}{ \dm{} $\downarrow$}\\
      \cmidrule(lr){2-12}
      & \multicolumn{11}{c}{MAE $\downarrow$} & & \\
    \midrule 
    \stl{}      & 0.07 & 0.18 &  60.6 &  53.9 & 0.50  &  4.53  &  58.8  &  64.2 &  63.8 &  66.2 & 0.07 & -    & -      \\
    \midrule
    \mgda{}     & 0.22 & 0.37 & 126.8 & 104.6 & 3.23  &  5.69  &  88.4  &  89.4 &  89.3 &  88.0 & 0.12 &7.73 & 120.5 \\
    \pcgrad{}   & 0.11 & 0.29 &  75.9 &  88.3 & 3.94  &  9.15  & 116.4  & 116.8 & 117.2 & 114.5 & 0.11 &6.09 & 125.7 \\
    \cagrad{}   & 0.12 & 0.32 &  83.5 &  94.8 & 3.22  &  6.93  & 114.0  & 114.3 & 114.5 & 112.3 & 0.12 & 7.09 & 112.8 \\
    \imtlg{}    & 0.14 & 0.29 &  98.3 &  93.9 & 1.75  &  5.70  & 101.4  & 102.4 & 102.0 & 100.1 & 0.10 & 5.91 &  77.2 \\
    \nashmtl{}  & \best{0.10} & \best{0.25} &  82.9 &  \best{81.9} & 2.43  &  5.38  &  74.5  &  75.0 &  75.1 &  74.2 & \best{0.09} &\best{3.64} &  62.0 \\
    \midrule
    \ls{}       & 0.11 & 0.33 &  \best{73.6} &  89.7 & 5.20  & 14.06  & 143.4  & 144.2 & 144.6 & 140.3 & 0.13 & 8.00 & 177.6 \\
    \si{}       & 0.31 & 0.35 & 149.8 & 135.7 & \best{1.00}  &  \best{4.51}  &  55.3  &  55.8 & 55.8 & 55.3 & 0.11 & 5.09 &  77.8 \\
    \rlw{}      & 0.11 & 0.34 &  76.9 &  92.8 & 5.87  & 15.47  & 156.3  & 157.1 & 157.6 & 153.0 & 0.14 & 9.36 & 203.8 \\
    \dwa{}      & 0.11 & 0.33 &  74.1 &  90.6 & 5.09  & 13.99  & 142.3  & 143.0 & 143.4 & 139.3 & 0.13 & 7.64 & 175.3 \\
    \uw{}       & 0.39 & 0.43 & 166.2 & 155.8 & 1.07  &  4.99  &  66.4  &  66.8 &  66.8 &  66.2 & 0.12 & 6.64 & 108.0 \\
    \famo{}     & 0.15 & 0.30 & 94.0 & 95.2 &  1.63 & 4.95 & 70.82 & 71.2 & 71.2 & 70.3 & 0.10 & 4.73 & 58.5 \\
    \ours{} &0.17	&0.35	&102.4	&119.0	&1.22	&4.94	&\best{53.9}	&\best{54.3}	&\best{54.3}	&\best{53.9}	&0.11	&4.55 &\best{52.7}
    \\
    \bottomrule 
    \end{tabular}
    }
    \label{tab: qm9_detail}
    \vspace{-10pt}
\end{table*}

\begin{table*}[t!]
    \centering
    \caption{\textbf{Detailed results on }~\texttt{CityScapes}. Each experiment is repeated over 3 random seeds and the mean is reported. The best average result is marked in bold. \mr{} and \dm{} are the main metrics for MTL performance.}
    \resizebox{0.8\textwidth}{!}{%
    \begin{tabular}{lrrrrrr}
    \toprule
   \multirow{2}{*}{\textbf{Method}}      &  \multicolumn{2}{c}{Segmentation} & \multicolumn{2}{c}{Depth} & \multirow{2}{*}{\mr{} $\downarrow$} & \multirow{2}{*}{ \dm{} $\downarrow$}  \\
    \cmidrule(lr){2-3}\cmidrule(lr){4-5}
     &  mIoU $\uparrow$ & Pix Acc $\uparrow$ & Abs Err $\downarrow$ & Rel Err $\downarrow$ &  & \\
    \midrule 
    \stl{}      & 74.01 & 93.16 & 0.0125 & 27.77 & & \\
    \midrule
    \mgda{}     & 68.84 & 91.54 & 0.0309 & 33.50 & 10.00 & 44.14  \\
    \pcgrad{}   & 75.13 & 93.48 & 0.0154 & 42.07 & 6.25 & 18.29  \\
    \cagrad{}   & 75.16 & 93.48 & 0.0141 & 37.60 & 5.00 & 11.64  \\
    \imtlg{}    & 75.33 & 93.49 & 0.0135 & 38.41 & 4.00 & 11.10  \\
    \nashmtl{}  & \best{75.41} & \best{93.66} & \best{0.0129} & 35.02 & \best{2.50} & 6.82  \\
    \midrule
    \ls{}       & 70.95 & 91.73 & 0.0161 & 33.83 & 8.50 & 14.11 \\
    \si{}       & 70.95 & 91.73 & 0.0161 & 33.83 & 8.50 & 14.11 \\
    \rlw{}      & 74.57 & 93.41 & 0.0158 & 47.79 & 7.75 & 24.38 \\
    \dwa{}      & 75.24 & 93.52 & 0.0160 & 44.37 & 6.00 & 21.45 \\
    \uw{}       & 72.02 & 92.85 & 0.0140 & 30.13 & 5.75 & \best{5.89} \\
    \famo{}     & 74.54 & 93.29 & 0.0145 & 32.59 & 5.50  & 8.13 \\
    \ours{} &72.63 &93.03 &0.0164 &\best{27.58} & 7.00 &8.11 \\
    \bottomrule 
    \end{tabular}
    }
    \label{tab: cityscapes_detail}
\end{table*}

%% file: table/time.tex
\begin{table}[t!]
    \centering
    \caption{\textbf{Detailed training time (seconds) for one epoch of different methods.} In addition to LS, our method requires the lowest time cost during training on all 4 datasets.}
    \resizebox{0.6\textwidth}{!}{%
    \begin{tabular}{lrrrr}
    \toprule
   {\textbf{Method}} & \textbf{NYU-v2} & {\textbf{QM9}} & {\textbf{CityScapes}} & {\textbf{CelebA}} \\
    \midrule
    \mgda{}     & 199.73 & 601.52  & 136.69 & 902.53 \\
    \pcgrad{}   & 192.41 & 352.05 & 100.21 & 1486.71  \\
    \cagrad{}   & 180.37 & 296.36  & 107.84 & 590.05 \\
    \imtlg{}    & 180.67 &  306.11 & 113.51 & 522.22  \\
    \nashmtl{}  & 201.30 &  585.37 & 138.26 & 2134.75  \\
    \midrule
    \ls{}       & 86.21 & 88.95 & 66.63 & 170.81 \\
    \famo{}     & 102.87 & 100.40 & 78.70  & 183.46 \\
    \ours{}     & 87.78 & 93.50 & 67.30 & 171.69 \\
    \bottomrule 
    \end{tabular}
    }
    \vspace{-10pt}
    \label{tab: trainingtime}
\end{table}


%% file: section/7_checklist.tex
\newpage
\section*{NeurIPS Paper Checklist}

\begin{enumerate}

\item {\bf Claims}
    \item[] Question: Do the main claims made in the abstract and introduction accurately reflect the paper's contributions and scope?
    \item[] Answer: \answerYes{} 
    \item[] Justification: 
    The contributions and scope of this paper are claimed in the abstract. Detailed information can be found in the introduction section~\ref{sec: introduction}.

\item {\bf Limitations}
    \item[] Question: Does the paper discuss the limitations of the work performed by the authors?
    \item[] Answer: \answerYes{}
    \item[] Justification: 
    We provide a "limitation" subsection in the conclusion section~\ref{sec: conclusion}.

\item {\bf Theory Assumptions and Proofs}
    \item[] Question: For each theoretical result, does the paper provide the full set of assumptions and a complete (and correct) proof?
    \item[] Answer: \answerNA{}
    \item[] Justification: 
    This paper does not include theoretical results.

    \item {\bf Experimental Result Reproducibility}
    \item[] Question: Does the paper fully disclose all the information needed to reproduce the main experimental results of the paper to the extent that it affects the main claims and/or conclusions of the paper (regardless of whether the code and data are provided or not)?
    \item[] Answer: \answerYes{}
    \item[] Justification: 
    We provide all the details of the experiment in the experiment section. The algorithm and Python codes of our method can be found in Appendix~\ref{sec: algorithm}.

\item {\bf Open access to data and code}
    \item[] Question: Does the paper provide open access to the data and code, with sufficient instructions to faithfully reproduce the main experimental results, as described in supplemental material?
    \item[] Answer: 
    \answerYes{}
    \item[] Justification: 
    We provide all experimental details in the experiment section. The algorithm and Python codes of our method can be found in Appendix~\ref{sec: algorithm}. Details about MTL benchmarks in this paper are provided in Appendix~\ref{sec: set_up}.

\item {\bf Experimental Setting/Details}
    \item[] Question: Does the paper specify all the training and test details (e.g., data splits, hyperparameters, how they were chosen, type of optimizer, etc.) necessary to understand the results?
    \item[] Answer: 
    \answerYes{}
    \item[] Justification: 
    We follow the standard experimental setup in MTO, where the data splits, hyperparameters, and optimizer are set as the same as previous works (nashMTL, CAGrad, and FAMO). Detailed information can be found in Appendix~\ref{sec: appendix/trainingdetails}.

\item {\bf Experiment Statistical Significance}
    \item[] Question: Does the paper report error bars suitably and correctly defined or other appropriate information about the statistical significance of the experiments?
    \item[] Answer: \answerYes{}
    \item[] Justification: 
     Following the standard experimental setup, we repeat each experiment over 3 random seeds and report the mean of the results.

\item {\bf Experiments Compute Resources}
    \item[] Question: For each experiment, does the paper provide sufficient information on the computer resources (type of compute workers, memory, time of execution) needed to reproduce the experiments?
    \item[] Answer: \answerYes{}
    \item[] Justification: 
    We provide the computing resources in Appendix~\ref{sec: appendix/trainingdetails}. The training-time efficiency is provided in Figure~\ref{fig: efficiency comparisons}.
    
\item {\bf Code Of Ethics}
    \item[] Question: Does the research conducted in the paper conform, in every respect, with the NeurIPS Code of Ethics \url{https://neurips.cc/public/EthicsGuidelines}?
    \item[] Answer: \answerYes{}
    \item[] Justification: 
    We reviewed and followed the NeurIPS Code of Ethics.

\item {\bf Broader Impacts}
    \item[] Question: Does the paper discuss both potential positive societal impacts and negative societal impacts of the work performed?
    \item[] Answer: \answerYes{}
    \item[] Justification: 
    We provide the potential broader impacts in the conclusion section~\ref{sec: conclusion}.
    
\item {\bf Safeguards}
    \item[] Question: Does the paper describe safeguards that have been put in place for responsible release of data or models that have a high risk for misuse (e.g., pretrained language models, image generators, or scraped datasets)?
    \item[] Answer: \answerNA{}
    \item[] Justification: The data and models pose no such risks.

\item {\bf Licenses for existing assets}
    \item[] Question: Are the creators or original owners of assets (e.g., code, data, models), used in the paper, properly credited and are the license and terms of use explicitly mentioned and properly respected?
    \item[] Answer: \answerYes{}
    \item[] Justification:
    We cite the original papers that produced the code package and datasets.

\item {\bf New Assets}
    \item[] Question: Are new assets introduced in the paper well documented and is the documentation provided alongside the assets?
    \item[] Answer: \answerYes{}
    \item[] Justification: The provided Python code cannot be used without the authors' permission.

\item {\bf Crowdsourcing and Research with Human Subjects}
    \item[] Question: For crowdsourcing experiments and research with human subjects, does the paper include the full text of instructions given to participants and screenshots, if applicable, as well as details about compensation (if any)? 
    \item[] Answer: \answerNA{}
    \item[] Justification: 
    This paper does not involve crowdsourcing nor research with human subjects.

\item {\bf Institutional Review Board (IRB) Approvals or Equivalent for Research with Human Subjects}
    \item[] Question: Does the paper describe potential risks incurred by study participants, whether such risks were disclosed to the subjects, and whether Institutional Review Board (IRB) approvals (or an equivalent approval/review based on the requirements of your country or institution) were obtained?
    \item[] Answer: \answerNA{}
    \item[] Justification: 
    This paper does not involve crowdsourcing nor research with human subjects.

\end{enumerate}